\PassOptionsToPackage{table}{xcolor}
\documentclass[sigconf, nonacm]{acmart}
\usepackage{enumitem}
\usepackage{array}
\usepackage[table]{xcolor}
\usepackage{multirow}
\usepackage{caption}
\usepackage{flushend}
\usepackage{hyperref}
\usepackage{subcaption}
\usepackage{balance}
\usepackage{setspace}
\usepackage{cleveref}
\usepackage{graphicx}
\usepackage{makecell}
\newcommand\vldbdoi{XX.XX/XXX.XX}
\newcommand\vldbpages{XXX-XXX}
\newcommand\vldbvolume{14}
\newcommand\vldbissue{1}
\newcommand\vldbyear{2020}
\newcommand\vldbauthors{\authors}
\newcommand\vldbtitle{\shorttitle} 
\newcommand\vldbavailabilityurl{URL_TO_YOUR_ARTIFACTS}
\newcommand\vldbpagestyle{plain} 
\begin{document}
\title{OpenGU: A Comprehensive Benchmark for Graph Unlearning}

\author{Bowen Fan}
\affiliation{%
  \institution{Beijing Institute of Technology}
  \streetaddress{}
  \city{}
  \state{}
  \postcode{}
}
\email{fan1085165825@gmail.com}

\author{Yuming Ai}
\orcid{}
\affiliation{%
  \institution{Beijing Institute of Technology}
  \streetaddress{}
  \city{}
  \country{}
}
\email{1120211787@bit.edu.cn}

\author{Xunkai Li}
\orcid{}
\affiliation{%
  \institution{Beijing Institute of Technology}
  \city{}
  \country{}
}
\email{cs.xunkai.li@gmail.com}

\author{Zhilin Guo}
\affiliation{%
  \institution{Shandong University}
  \city{}
  \country{}
}
\email{frank04180@outlook.com}

\author{Rong-Hua Li}
\affiliation{%
  \institution{Beijing Institute of Technology}
  \city{}
  \country{}
}
\email{lironghuabit@126.com}

\author{Guoren Wang}
\affiliation{%
  \institution{Beijing Institute of Technology}
  \city{}
  \country{}
}
\email{wanggrbit@gmail.com}

\begin{abstract}

Graph Machine Learning is essential for understanding and analyzing relational data. However, privacy-sensitive applications demand the ability to efficiently remove sensitive information from trained graph neural networks (GNNs), avoiding the unnecessary time and space overhead caused by retraining models from scratch.
To address this issue, Graph Unlearning (GU) has emerged as a critical solution, with the potential to support dynamic graph updates in data management systems and enable scalable unlearning in distributed data systems while ensuring privacy compliance.
Unlike machine unlearning in computer vision or other fields, GU faces unique difficulties due to the non-Euclidean nature of graph data and the recursive message-passing mechanism of GNNs. Additionally, the diversity of downstream tasks and the complexity of unlearning requests further amplify these challenges.
Despite the proliferation of diverse GU strategies, the absence of a benchmark providing fair comparisons for GU, and the limited flexibility in combining downstream tasks and unlearning requests, have yielded inconsistencies in evaluations, hindering the development of this domain. To fill this gap, we present OpenGU, the first GU benchmark, where 16 SOTA GU algorithms and 37 multi-domain datasets are integrated, enabling various downstream tasks with 13 GNN backbones when responding to flexible unlearning requests. 
 Based on this unified benchmark framework, we are able to provide a comprehensive and fair evaluation for GU. Through extensive experimentation, we have drawn $8$ crucial conclusions about existing GU methods, while also gaining valuable insights into their limitations, shedding light on potential avenues for future research.
\end{abstract}

\maketitle
\pagestyle{\vldbpagestyle}
\begingroup\small\noindent\raggedright\textbf{PVLDB Reference Format:}\\
\vldbauthors. \vldbtitle. PVLDB, \vldbvolume(\vldbissue): \vldbpages, \vldbyear.\\
\href{https://doi.org/\vldbdoi}{doi:\vldbdoi}
\endgroup
\begingroup
\renewcommand\thefootnote{}\footnote{\noindent
This work is licensed under the Creative Commons BY-NC-ND 4.0 International License. Visit \url{https://creativecommons.org/licenses/by-nc-nd/4.0/} to view a copy of this license. For any use beyond those covered by this license, obtain permission by emailing \href{mailto:info@vldb.org}{info@vldb.org}. Copyright is held by the owner/author(s). Publication rights licensed to the VLDB Endowment. \\
\raggedright Proceedings of the VLDB Endowment, Vol. \vldbvolume, No. \vldbissue\ %
ISSN 2150-8097. \\
\href{https://doi.org/\vldbdoi}{doi:\vldbdoi} \\
}\addtocounter{footnote}{-1}\endgroup

\ifdefempty{\vldbavailabilityurl}{}{
\vspace{.3cm}
\begingroup\small\noindent\raggedright\textbf{PVLDB Artifact Availability:}\\
The source code, data, and/or other artifacts have been made available at \url{https://github.com/bwfan-bit/OpenGU}.
\endgroup
}
\section{Introduction}
Graphs are versatile mathematical structures that represent complex interactions and relationships between entities, offering an abstract yet intuitive framework for modeling real-world systems. In the context of database systems, graphs provide a powerful means to represent and analyze relational data, enabling insights into structured information. To effectively capture the rich and interconnected information inherent in graph data, GNNs \cite{2016Convolutional_ChebNet,sun2023adpa,luo2024classic} have emerged as transformative tools, achieving remarkable success in optimizing database query performance \cite{qgdatabase,robinson2015graph,ngdatabase,cvitkovic2020supervised}, enhancing data management \cite{angles2018introduction,aggarwal2010graph}, and supporting other fields such as social networks \cite{app10041327,ijcai2018p142}, recommendation systems \cite{su2024dcl, LightGCN, cai2023app_gnn_rec3}, biological networks \cite{NIPS2017_f5077839,10.1093/bioinformatics/bty294,qu2023app_gnn_bio2} and data mining \cite{ieeedataming2012,wsdmdataming2021}.


Nowadays, the majority of machine learning paradigms are primarily driven by data for the acquisition of knowledge, fundamentally transforming decision-making processes across various fields \cite{wu2020gnn_survey1,zhou2022gnn_survey2,bessadok2022gnn_survey3}. However, the indiscriminate use of data has intensified the conflict between data rights and user privacy \cite{liu2020privacy,tanuwidjaja2020privacy,wu2022survey}. In response to these pressing issues, a range of pioneering regulatory frameworks has been proposed, including the European Union’s General Data Protection Regulation (GDPR) \cite{GDPR}, and the California Consumer Privacy Act (CCPA) \cite{CCPA1} in California.
Among these regulations, one of the most critical and contentious provisions is \emph{the right to be forgotten} \cite{kwak2017let}. To technically align the effectiveness of machine learning with data rights and privacy regulations, a revolutionary frontier—machine unlearning \cite{Cao2015Towards, Xu2023ACMMU} has emerged.

Despite advances in traditional machine unlearning for independently distributed data, the growing reliance on graph-structured data in various fields like database systems and data mining highlights the significance of GU. As graphs underpin relational schemas and mined patterns, GU enables selective removal of sensitive information while preserving the integrity of analytical results. 
Unlike other domains, removing specific information from a graph requires not only deleting or modifying individual nodes or edges but also needs to consider how these changes ripple through the entire graph. In addition, GU is jointly determined by downstream tasks and unlearning requests. Downstream tasks include node classification, link prediction, and graph classification (See Sec. \ref{downstream}), while unlearning requests span multiple levels, such as node, edge, and feature (See Sec. \ref{unlearning_request}). The interplay of these two aspects makes GU even more complex.
These challenges position GU as a particularly demanding area of research, requiring tailored solutions for its structural complexities and task-specific needs.
Recently, a variety of GU methods has surfaced, paving the way for more responsible and privacy-compliant graph learning through data-level techniques, architectural changes, and parameter update strategies.

Though these GU strategies have made strides from various perspectives, there are still deficiencies in achieving unity: (1) \textbf{Datasets}: The utilization of different datasets (i.e., domain and scale) and rigidity of processing data regarding splitting and inference settings make it difficult to analyze the results. 
(2) \textbf{Backbones}: The lack of harmonization among GNN backbones between different methods creates barriers for fair comparisons. (3) \textbf{Experiments}: Typically, experiments are conducted on a single downstream task under one type of unlearning request, which prevents cross-examining of downstream tasks under different unlearning requests. Besides, the varying configurations of unlearning requests and differing evaluation metrics in existing methods lead to divergent interpretations in experimental results. These challenges highlight the urgent need for a standardized benchmark to establish a unified and comprehensive understanding of the GU landscape.

To address the aforementioned challenges, we propose OpenGU in this paper, to the best of our knowledge, the first comprehensive benchmark specifically designed for graph unlearning. OpenGU integrates $16$ recently proposed SOTA algorithms and $37$ multi-domain datasets, enabling the flexible $3\times3$ combination of unlearning requests and downstream tasks. OpenGU implements these components with unified APIs to address three types of unlearning requests. Based on this design, we conduct an in-depth investigation of these GU algorithms, offering valuable insights in terms of three dimensions: effectiveness, efficiency, and robustness. For \textbf{effectiveness}, OpenGU provides a comprehensive evaluation in two aspects: model updating and inference protection. We fairly assess the forgetting ability for unlearning entities and the reasoning capability for retained data to determine whether the methods achieve effective trade-off between forgetting and reasoning. For \textbf{efficiency}, OpenGU provides insights into scalability by evaluating how well GU methods can handle increasing amounts of data and more complex graph structures, showing their practical applicability in real-world scenarios. Furthermore, we analyze the computational resources required by various graph unlearning algorithms, focusing on time and space complexity from both theoretical and empirical perspectives. For \textbf{robustness}, we focus on noise and sparsity scenarios in real-world applications, and delve into the algorithms' capacity to maintain performance stability and integrity amid different unlearning intensities. 

\textbf{Our contributions}. We propose OpenGU, the inaugural benchmark specifically tailored for GU domain. Our contributions are outlined as follows:
(1) \underline{\textit{Comprehensive Benchmark.}} OpenGU integrates $16$ SOTA algorithms and $37$ popular multi-domain datasets, establishing a thorough evaluation framework for a fair comparison. Moreover, OpenGU expands and standardizes experimental task requirements, encompassing unlearning requests and downstream tasks, thereby facilitating a code-level implementation of $3\times3$ cross-experiment configurations.
(2) \underline{\textit{Analytical Insights.}} Through extensive empirical experiments centered around algorithms' forgetting capability and reasoning capability, we conduct a thorough analysis 
and summarize $8$ key conclusions, envisioning our conclusions as the catalyst for GU field. 
(3) \underline{\textit{Open-sourced Benchmark Library.}} OpenGU is designed as an open-source benchmark library, providing researchers and practitioners with accessible tools and resources for expanding the exploration of GU. Additionally, comprehensive documentation and user-friendly interfaces foster collaboration and innovation within the GU community. Our code is available at \href{https://github.com/bwfan-bit/OpenGU}{https://github.com/bwfan-bit/OpenGU}.

\captionsetup[figure]{}
\begin{figure*}[h]
  \centering
  \includegraphics[width=1\linewidth,height=0.55\linewidth]{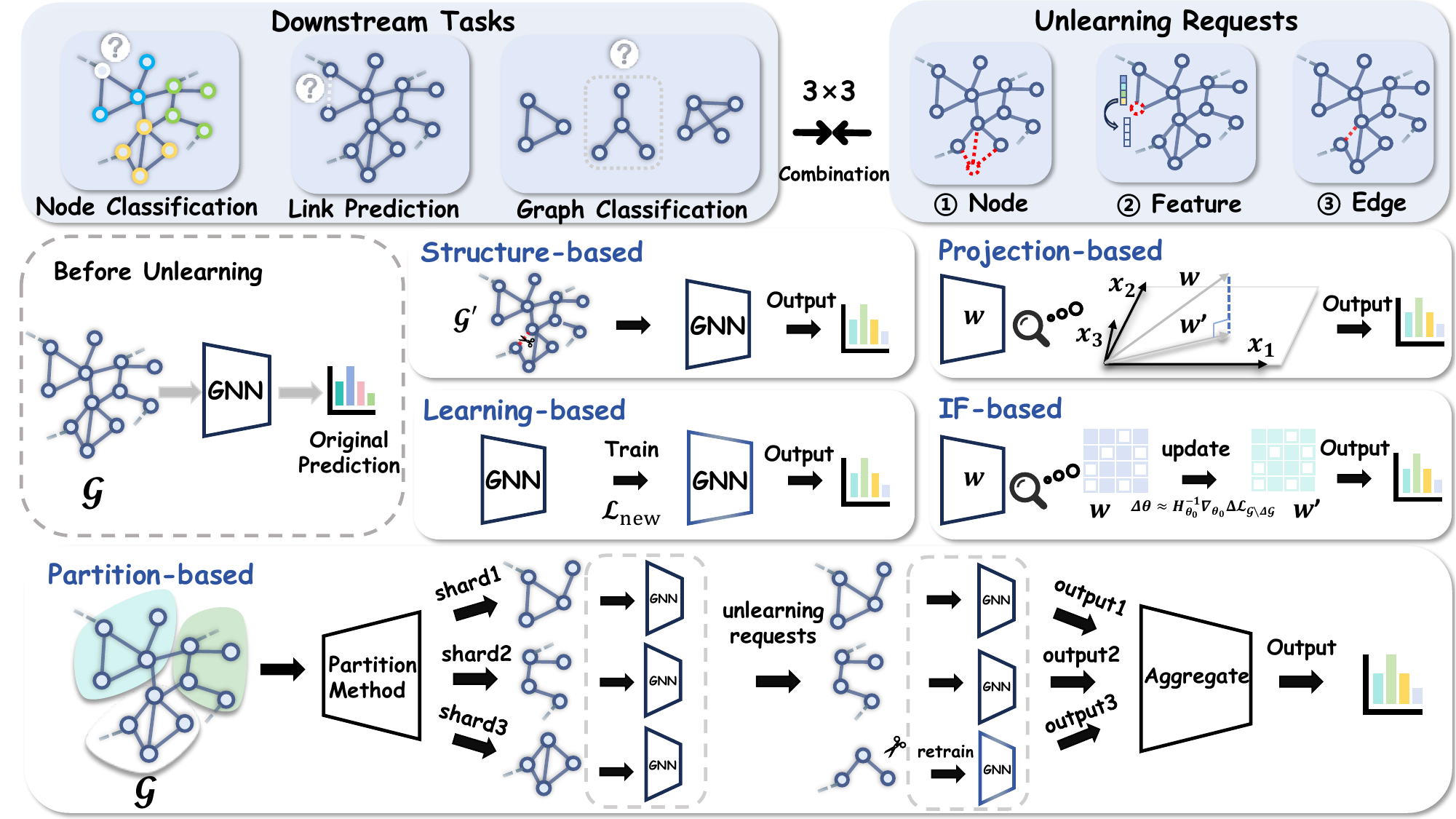}
  \caption{An overview of the OpenGU framework, illustrating the key components and methodologies involved in GU.}
  \vspace{-0.3cm}
  \label{figure1}
\end{figure*}

\section{Definitions and Background}

In this section, we will briefly review several key concepts and definitions to better explain the fundamentals of GU. Generally, we define a graph as $\mathcal{G}=(\mathcal{V},\mathcal{E},\mathcal{X})$, where $\mathcal{V}$ represents a set of nodes with $\lvert \mathcal{V} \rvert=n$, and $\mathcal{E}$ denotes the edge set containing $\lvert \mathcal{E} \rvert = m$ edges. The feature matrix, $\mathcal{X} \in\mathbb{R}^{n\times d}$, represents the feature vectors of all nodes, and $d$ represents the dimension of node features. We also define $\mathcal{Y} = \{y_1,y_2,...,y_n\}$ as the label set, where each $y_i$ corresponds to the node $v_i \in \mathcal{V}$ in a one-to-one manner. Apart from the aforementioned attributes, the graph is also characterized by its adjacency matrix $A \in \mathbb{R}^{n\times n}$, where each element $A_{ij}$ indicates the edge between nodes $i$ and $j$. Formally, we define $\mathcal{M}$ as the original model trained on a complete graph $\mathcal{G}$ or a graph dataset denoted as $G = \{\mathcal{G}_1,\mathcal{G}_2,...,\mathcal{G}_z\}$ with the corresponding labels $L = \{l_1,l_2,...,l_z\}$.

\subsection{Graph Neural Networks}
In this part, we mainly focus on Message Passing Neural Networks, a widely adopted paradigm within GNNs for learning graph data. These models leverage the structural properties of graphs to learn node representations effectively. Broadly, such GNNs can be defined through three key components: initialization, aggregation, and update, offering a unified framework to capture relational dependencies in graphs.

\noindent\textbf{1.Initialization.}
To initialize node representations in a graph, each node is typically assigned an embedding based on its features or attributes. This process can be formalized as:
\begin{equation}
h_v = x_v\quad\forall v \in \mathcal{V}.
\end{equation}

\noindent\textbf{2.Aggregation.}
To capture information from the neighborhood of node $u$, an aggregation function $f_{Agg}^{(k-1)}$ is applied to combine the embeddings of $\mathcal{N}(u)$ from the previous iteration. 
Common aggregators include permutation-invariant functions like sum, mean, or max, ensuring that the aggregated result is independent of the order of neighbors. This operation can be expressed as:
\begin{equation}
m_u^{(k)} = f_{Agg}^{(k-1)}(\{h_v^{(k-1)},v\in \mathcal{N}(u)\}).
\end{equation}
\noindent\textbf{3.Update.}
This process enables each node to refine its representation using $f_{Up}^{(k-1)}$, which combines the node's embedding and the neighborhood's aggregated message.
Specifically, the updated embedding for node $u$ is computed as follows:
\begin{equation}
h_u^{(k)} = f_{Up}^{(k-1)}(h_v^{(k-1)},m_{\mathcal{N}(u)}^{(k-1)}).
\end{equation}

The combination of initialization, aggregation, and update functions forms the foundation of GNNs. Through iterative refinement, node embeddings progressively incorporate structural and feature information from their local neighborhoods, enabling GNNs to learn rich and expressive representations for downstream tasks.

\subsection{Downstream Tasks}
\label{downstream}
Building upon the iterative process of node representation, the resulting node embeddings $h_u^{(k)}$ at the final iteration $k$ serve as the foundation for downstream tasks. These refined embeddings are rich in information and capture both local and global graph structures, making them highly suitable for various graph-based prediction tasks. The downstream tasks typically involve node classification, link prediction, and graph classification, each utilizing these node embeddings in different ways.

\textit{Node classification} aims to assign labels to individual nodes in the graph based on their features and structural context. The learned node representations $h^{(k)}$ are fed into a classifier, often a fully connected layer, to predict node labels. For instance, in financial transaction networks, this task can help determine whether a given transaction is fraudulent or not. 

\textit{Link prediction} seeks to predict the existence of edges between nodes using their embeddings and graph structure. This is critical in tasks like drug-drug interaction prediction in biomedical networks, where nodes represent drugs, and edges indicate known interactions. Predicting potential edges can aid in discovering new therapeutic drug combinations. 

\textit{Graph classification} predicts a label for an entire graph by aggregating node embeddings $h^{(k)}$ via a readout function $f_{Read}$ (e.g., sum or mean). In chemical compound classification, each molecule can be represented as a graph, where nodes correspond to atoms, and edges represent chemical bonds. Graph classification can help predict whether a molecule is toxic or not.

\definecolor{myblue1}{RGB}{144, 201, 230}
\definecolor{myblue2}{RGB}{155, 187, 225}
\definecolor{myblue3}{RGB}{122, 187, 219}
\definecolor{myblue}{RGB}{146, 220, 247}
\definecolor{mycyan}{RGB}{151, 251, 241}
\definecolor{mygreen}{RGB}{220, 254, 141}
\definecolor{mygreen2}{RGB}{237, 254, 198}
\definecolor{myblue4}{RGB}{196, 225, 255}
\definecolor{myblue5}{RGB}{36,80,138}
\definecolor{under}{RGB}{116,141,164}
\definecolor{mygray}{gray}{.9}
\captionsetup[table]{}
\begin{table*}[t]
\caption{An overview of OpenGU.}
\renewcommand{\aboverulesep}{0.7pt}
\renewcommand{\belowrulesep}{0.7pt}
\label{table1}
\centering 
\fontsize{5pt}{6pt}\selectfont
\renewcommand{\arraystretch}{1} 
\resizebox{0.9\textwidth}{!}{
\begin{tabular}{>{\centering\arraybackslash}p{2cm}|
    >{\centering\arraybackslash}p{6cm}}
\midrule[0.08em]
\rowcolor{white} 
\multicolumn{2}{c}{\cellcolor{mygray} \emph{\textbf{Datasets}}}  \\ \midrule[0.08em]

{Type} & {Homophily, Heterophily, Node, Edge, Graph}\\  
{Domain} & {Citation, Co-author, Social, Wiki, Image, Protein, Movie, ...} \\
Preprocess & { Transductive/Inductive, Label-Balanced/Label-Random, Noise, Sparsity}\\
\midrule [0.08em]
\rowcolor{white}
\multicolumn{2}{c}{\cellcolor{mygray}\emph{\textbf{GNN Algorithms}}} \\ \midrule[0.08em]
{Decoupled}& {SGC, SSGC, SIGN, APPNP} \\
{Sampling} &{GraphSAGE, GraphSAINT, Cluster-gcn} \\ 

{Traditional} &{GCN, GCNII, LightGCN, GAT, GATv2, GIN} \\ 
\midrule[0.08em]
\multicolumn{2}{c}{\cellcolor{mygray}\emph{\textbf{ Mainstream GU Algorithms}}} \\ \midrule[0.08em]
{Partition-based} &  {GraphEraser, GUIDE, GraphRevoker} \\ 
{IF-based} & {GIF, CGU, CEU, GST, IDEA,  ScaleGUN} \\
{Learning-based} & {GNNDelete, MEGU, SGU, D2DGN, GUKD} \\ 
\midrule[0.08em]
\rowcolor{white}
\multicolumn{2}{c}{\cellcolor{mygray}\emph{\textbf{Evaluations}}} \\ \midrule[0.08em]

{Attack}  & {Membership Inference Attack, Poisoning Attack}\\
{Effectiveness}  & {Accuracy, Precision, F1-score, AUC-ROC} \\ 
Efficiency & {Time, Memory, Theoretical Algorithm Complexity} \\ 

Robustness & {Deletion Intensity, GU Scenario Noise and Sparsity} \\
Unlearning Request &  {Node-level Request, Feature-level Request, Edge-level Request} \\
Downstream Task & {Node Classification, Link Prediction, Graph Classification}\\
 \midrule[0.08em] 
\end{tabular}}
\end{table*}

\subsection{Unlearning Requests}
\label{unlearning_request}
When receiving unlearning request $\Delta \mathcal{G}=\{\Delta \mathcal{V}, \Delta \mathcal{E}, \Delta \mathcal{X}\}$, the retrained model $\hat{\mathcal{M}}$ is trained from scratch on the pruned graph after the deletion and the unlearning model $\mathcal{M'}$ updates its parameters from original $W$ to $W'$ based on the unlearning algorithm. The goal of GU is to minimize the discrepancy between $\mathcal{M'}$ and $\hat{\mathcal{M}}$. Common unlearning requests in graph unlearning typically fall into three categories: node-level $\Delta \mathcal{G}=\{\Delta \mathcal{V}, \varnothing, \varnothing\}$, edge-level  $\Delta \mathcal{G}=\{\varnothing, \Delta \mathcal{E}, \varnothing\}$, and feature-level $\Delta \mathcal{G}=\{\varnothing, \varnothing, \Delta \mathcal{X}\}$. 

Node unlearning targets the removal of individual nodes, facilitating the precise deletion of specific data points, such as individual users or entities, while preserving the broader structure of the graph. In social network platforms, if users (nodes) decide to delete their accounts, node unlearning ensures that the model no longer uses their interactions or behaviors to influence recommendations or predictions, safeguarding user privacy. Edge unlearning, on the other hand, targets the removal of relationships between nodes—an approach essential for privacy-sensitive applications where certain connections require erasure without altering node attributes. A key application of edge unlearning is in financial fraud detection, where certain transactional relationships (edges) between users may need to be erased if they are identified as erroneous or compromised. Feature unlearning, meanwhile, involves the elimination of specific node features, facilitating controlled deletion of attribute-based information linked to individual nodes. For example, in healthcare, sensitive medical attributes can be excluded to comply with regulations while retaining the utility of remaining features.

\vspace{-0.2cm}
\subsection{GU Taxonomy}
To provide a comprehensive understanding of GU, we categorize existing methodologies into five types based on their operational frameworks: (1) Partition-based algorithms, (2) Influence Function-based (IF-based) unlearning algorithms, (3) Learning-based algorithms, (4) Projection-based algorithms, and (5) Structure-based algorithms. In Partition-based methods, GraphEraser \cite{chen2022graph_eraser}, GUIDE \cite{wang2023guide}, and GraphRevoker \cite{2024ZhangGraphRevoker} draw inspiration from SISA \cite{2021SISA} to implement different partitioning strategies, enabling training and unlearning on independent shards. In IF-based methods, GIF \cite{wu2023gif}, CGU \cite{chien2022cgu}, CEU \cite{2023WuCEU}, and others \cite{pan2023gst_unlearning,2024IDEA,scaluGUN} leverage rigorous mathematical formulations to quantify the impact of data removal on model, allowing for efficient model updates. In Learning-based algorithms methods, GNNDelete \cite{cheng2023gnndelete}, MEGU \cite{xkliMEGU2024}, SGU \cite{}, and others \cite{2024D2DGN,2023GUKD,2023GCU} achieve a trade-off between forgetting and reasoning with specialized loss functions. As for Projection-based algorithm, Projector \cite{cong2023projector} adapts to unlearning by orthogonally projecting the weights into a different subspace. Finally, Structure-based method UtU \cite{Tan2024UtU} manipulate the graph structure directly without the extensive retraining or complex optimizations. This categorization provides a structured landscape of GU, as illustrated in Figure \ref{figure1}.

\section{Benchmark Design}

In this section, we present a comprehensive overview of OpenGU, emphasizing the key aspects of the benchmark design, as outlined in Table \ref{table1}. First, we detail the datasets and the associated preprocessing techniques (Sec. \ref{dataset}), followed by an examination of the various GU methods (Sec. \ref{alg}). Lastly, we outline the evaluation metrics and experimental configurations (Sec. \ref{eva}) that structure the benchmarking process, offering the core principles and the holistic view of our OpenGU framework.

\captionsetup[table]{}
\begin{table*}[t]
\caption{Statistical Overview of Datasets for Node and Edge-Level Tasks in OpenGU Benchmark.}
\renewcommand{\aboverulesep}{0.8pt}
\renewcommand{\belowrulesep}{0.8pt}
\renewcommand{\arraystretch}{0.9}
\fontsize{3pt}{4pt}\selectfont
\label{table2}
\centering 
\resizebox{0.95\textwidth}{!}{
\begin{tabular}{ccccccc}
    \arrayrulecolor{myblue5}\toprule[0.16em]
    \textbf{Datasets}  & \textbf{Nodes} & \textbf{Edges} & \textbf{Features } & \textbf{Classes} & \textbf{Type} & \textbf{Description}\\
    \arrayrulecolor{under}\midrule[0.1em]
    Cora & 2,708 & 5,278 & 1,433 & 7 & Homophily & Citation Network\\
    Citeseer & 3,327 & 4,732 & 3,703 & 6 & Homophily& Citation Network\\
    PubMed & 19,717 & 44,338 & 500 & 3 & Homophily& Citation Network\\ 
    DBLP & 17,716 & 52,867 & 1,639 & 4 & Heterophily & Citation Network\\
    ogbn-arxiv & 169,343 & 1,166,243 & 128 & 40 & Homophily & Citation Network\\ \arrayrulecolor{myblue4}\midrule[0.08em] \addlinespace[-0.75pt]
\arrayrulecolor{under} \midrule[0.08em]
    CS & 18,333 & 81,894 & 6,805 & 15 & Homophily& Co-author Network\\
    Physics & 34,493 & 247,962 & 8,415 & 5 & Homophily& Co-author Network\\ \arrayrulecolor{myblue4}\midrule[0.08em] \addlinespace[-0.7pt]
\arrayrulecolor{under} \midrule[0.08em]
    Photo & 7,487 & 119,043 & 745 & 8 & Homophily& Co-purchasing Network\\ 
    Computers & 13,381 & 245,778 & 767 & 10 & Homophily & Co-purchasing Network\\ 
    ogbn-products & 2,449,029 & 61,859,140 & 100 & 47 & Homophily & Co-purchasing Network\\ \arrayrulecolor{myblue4}\midrule[0.08em] \addlinespace[-0.7pt]
\arrayrulecolor{under} \midrule[0.08em]
    Chameleon & 2,277 & 36,101 & 2,325 & 5 & Heterophily  & Wiki-page Network \\
    Squirrel & 5,201 & 216,933 & 2,089 & 5 & Heterophily  & Wiki-page Network \\ \arrayrulecolor{myblue4}\midrule[0.08em] \addlinespace[-0.75pt]
\arrayrulecolor{under} \midrule[0.08em]

    Actor & 7,600 & 29,926 & 931 & 5 & Heterophily  &  Actor Network \\
    Minesweeper & 10,000 & 39,402 & 7 & 2 & Homophily &  Game Synthetic Network\\
    Tolokers & 11,758 & 519,000 & 10 & 2 & Homophily &  Crowd-sourcing Network\\
    Roman-empire & 22,662 & 32,927 & 300 & 18 & Heterophily  &  Article Syntax Network\\
    Amazon-ratings & 24,492 & 93,050 & 300 & 5 & Heterophily  &  Rating Network\\
    Questions & 48,921 & 153,540 & 301 & 2 & Homophily &  Social Network\\
    Flickr & 89,250 & 899,756 & 500 & 7 & Heterophily & Image Network\\ 
        \arrayrulecolor{myblue5}\bottomrule[0.16em]
\end{tabular}}
\end{table*}

\captionsetup[table]{}
\begin{table*}[t]
\caption{Statistical Overview of Datasets for Graph-Level Tasks in OpenGU Benchmark.}
\label{table3}
\centering 
\renewcommand{\aboverulesep}{0.7pt}
\renewcommand{\belowrulesep}{0.7pt}
\fontsize{2.5pt}{3.5pt}\selectfont
\renewcommand{\arraystretch}{0.9} 
\resizebox{0.95\textwidth}{!}{
\begin{tabular}{ccccccc}
    \arrayrulecolor{myblue5}\toprule[0.16em]
    \textbf{Datasets}  & \textbf{Graphs} &\textbf{Nodes} & \textbf{Edges} & \textbf{Features } & \textbf{Classes} & \textbf{Description}\\
    \arrayrulecolor{under}\midrule[0.1em]
    MUTAG & 188 & 17.93 & 19.79 & 7 & 2 & Compounds Network\\
    PTC-MR & 344 & 14.29 & 14.69 & 18 & 2& Compounds Network\\
    BZR & 405 & 35.75 & 38.36 & 56 & 2 & Compounds Network\\ 
    COX2 & 467 & 41.22 & 43.45 & 38 & 2 & Compounds Network\\
    DHFR & 467 & 42.43 & 44.54 & 56 & 2 & Compounds Network\\ 
    AIDS & 2,000  & 15.69 & 16.20 & 42 & 2 & Compounds Network\\
    NCI1 & 4,110 & 29.87 & 32.30 & 37 & 2 & Compounds Network\\ 
    ogbg-molhiv & 41,127 & 25.50 & 27.50 & 9 & 2 & Compounds Network\\ 
    ogbg-molpcba & 437,929 & 26.00 & 28.10 & 9 & 2 & Compounds Network\\ \arrayrulecolor{myblue4}\midrule[0.08em] \addlinespace[-0.7pt]
\arrayrulecolor{under} \midrule[0.08em]
    ENZYMES & 600 & 32.63 & 62.14  & 21 & 6 & Protein Network\\ 
    DD & 1,178 & 284.32 & 715.66 & 89 & 2 & Protein Network\\ 
    PROTEINS & 1,113 & 39.06 & 72.82 & 4 & 2 & Protein Network\\ 
    ogbg-ppa & 158,100 & 243.40 & 2,266.10 & 4 & 37 & Protein Network\\ \arrayrulecolor{myblue4}\midrule[0.08em] \addlinespace[-0.7pt]
\arrayrulecolor{under} \midrule[0.08em]
    IMDB-BINARY &1,000 &19.77 &96.53 &degree &2 & Movie Network \\
    IMDB-MULTI& 1,500 &13.00 &65.94 &degree &3 &Movie Network \\  \arrayrulecolor{myblue4}\midrule[0.08em] \addlinespace[-0.7pt]
\arrayrulecolor{under} \midrule[0.08em]
    COLLAB &5,000 &74.49 &2,457.78 &degree &3  &Collaboration Network \\
    ShapeNet &16,881 &2,616.20 &KNN &3 &50 &Point Cloud Network \\ 
    MNISTSuperPixels &70,000 &75.00 &1,393.03 &1 &10 &Super-pixel Network \\
    \arrayrulecolor{myblue5}\bottomrule[0.16em]
\end{tabular}}
\vspace{-0.15cm}
\end{table*}

\vspace{-1em}
\subsection{Dataset Overview for OpenGU}
\label{dataset}

GU scenarios are fundamentally data-driven, making the meticulous selection of datasets indispensable for evaluating the GU strategies. To assess GU methods for node or edge-related tasks, we have carefully selected 19 datasets \cite{kipf2016gcn}. Citation Networks include Cora, Citeseer, and PubMed \cite{Yang16cora}, while Co-author Networks are represented by CS and Physics \cite{shchur2018amazon_datasets}. Image Networks feature Flickr \cite{zeng2019graphsaint}, while E-commerce and Product Networks incorporate Photo, Computers \cite{shchur2018amazon_datasets}, ogbn-products \cite{hu2020ogb}, and Amazon-ratings \cite{platonov2023hete_gnn_survey4}, capturing consumer and product interactions. Scientific Networks leverage DBLP \cite{2019DBLP} and ogbn-arxiv \cite{hu2020ogb} to reflect publication patterns. To further broaden the scope, we add Squirrel and Chameleon \cite{pei2020geomgcn}, reflecting webpage networks; Actor \cite{pei2020geomgcn}, focusing on film connections; and digital engagement data such as Minesweeper \cite{platonov2023hete_gnn_survey4} for online gaming, and Tolokers \cite{platonov2023hete_gnn_survey4}, from a crowdsourcing platform where edges signify task collaborations. Additionally, historical and social Q\&A contexts are represented by Roman-empire and Questions \cite{platonov2023hete_gnn_survey4}, respectively, enriching our dataset diversity.

Considering the scenario of graph classification tasks, OpenGU includes 18 datasets spanning various domains, Specifically, the compounds networks (MUTAG, PTC-MR, BZR, COX2, DHFR, AIDS, NCI1, ogbg-molhiv and ogbg-molpcba) \cite{MUTAG,PTC,BZR_COX_DHFR,AIDS,NCI1,hu2020ogb}, focus on molecular structures and chemical properties; the protein networks (ENZYMES, DD, PROTEINS and ogbg-ppa) \cite{chen2024fedgl,DD,PROTEINS,hu2020ogb} are concerned with biology data; the movie networks (IMDB-BINARY, IMDB-MULTI) \cite{IMDB} and collaboration networks (COLLAB) \cite{COLLAB} pertain to social media and community structures; additionally, ShapeNet \cite{ShapeNet} and MNISTSuperPixels \cite{MNISTSuperPixels} represent tasks related to 3D shapes and image superpixels. 
To provide a clearer understanding of the datasets, a detailed overview is showed in Tables \ref{table2} and \ref{table3} and further details about datasets can be found in \cite{OpenGU2025} (A.1).

In terms of data preprocessing, our work introduces several key enhancements to address existing limitations in splitting, inference, and special scenes with noise or sparsity,
making OpenGU more adaptable and comprehensive. GU methods currently lack a unified approach for dataset splitting: for instance, GraphEraser applies an 80\%/20\% training-test split, whereas GIF uses 90\%/10\%.
Although some datasets provide default splitting, these fixed ratios can limit the flexibility needed for diverse experimentation. 
To achieve a standardized and versatile splitting in OpenGU, we implemented code that allows arbitrary dataset split ratios, enabling researchers to customize splitting to suit their needs and experiment requirements.
Furthermore, we introduce a preprocessing enhancement that allows datasets to function under both transductive and inductive inference scenarios. While some datasets are typically optimized for only one inference scenario, this additional flexibility permits researchers to evaluate GU methods under both settings, thus offering a broader evaluation of algorithm performance across various inference contexts. Together, these contributions in data preprocessing make OpenGU a powerful benchmark framework for comprehensive and adaptable GU method assessment.

\subsection{Algorithm Framework for OpenGU}
\label{alg}
\textbf{GNN Backbones.} To evaluate the generalizability of GU algorithms, we incorporate three predominant paradigms of GNNs within our benchmark: traditional GNNs, sampling GNNs, and decoupled GNNs. For the traditional GNNs, we implement widely-recognized models such as GCN \cite{kipf2016gcn}, GAT \cite{velivckovic2017gat}, GIN \cite{xu2018gin} and others \cite{chen2020gcnii, brody2021gatv2, LightGCN}.
Sampling GNNs include GraphSAGE \cite{hamilton2017graphsage}, GraphSAINT \cite{zeng2019graphsaint} and Cluster-gcn \cite{chiang2019cluster-gcn}. Additionally, to accommodate GU methods which rely on linear-GNN, we further incorporate scalable, decoupled GNN models into the benchmark, specifically SGC \cite{wu2019sgc}, SSGC \cite{zhu2021ssgc}, SIGN \cite{frasca2020sign} and APPNP \cite{2019appnp}. These models offer scalability advantages and efficient decoupling for handling larger datasets and supporting diverse GU methods. 
More detailed descriptions of these GNN backbones are provided in \cite{OpenGU2025} (A.2).

\noindent \textbf{GU Algorithms.} For GU algorithms, our framework encompasses 16 methods, each meticulously reproduced based on source code or detailed descriptions in the relevant publications. 
The detailed descriptions can be found in \cite{OpenGU2025} (A.3). 
Moreover, we deliver a unified interface for GU methods, merging them under a cohesive API to facilitate easier access, experimentation, and future expansion. By standardizing these methods within OpenGU, we provide a streamlined and highly efficient platform for researchers and practitioners to conduct robust, reliable, and reproducible benchmarking studies. 

\vspace{-0.2cm}
\subsection{Evaluation Strategy for OpenGU}
\label{eva}

To assess GU algorithms in diverse real-world scenarios, our benchmark evaluation spans three critical dimensions tailored to GU contexts: effectiveness, efficiency, and robustness. Each dimension includes tailored evaluation methods reflecting OpenGU’s mission to serve as a flexible, high-standard benchmark.

 \textbf{Cross-over Design.}
In previous GU studies, node and feature unlearning typically align with node classification tasks, while edge unlearning is often evaluated in the context of link prediction. However, real-world applications frequently demand the removal of data in scenarios where unlearning requests and downstream tasks intersect. 
To address this gap, we designed cross-task evaluations in OpenGU, allowing us to measure GU algorithm performance in more complex, realistic scenarios where different unlearning types may apply across diverse downstream tasks. This approach provides a comprehensive and practical evaluation framework to assess the flexibility of GU algorithms in real-world applications.

\definecolor{myblue1}{RGB}{144, 201, 230}
\definecolor{myblue2}{RGB}{155, 187, 225}
\definecolor{myblue3}{RGB}{122, 187, 219}
\definecolor{myblue}{RGB}{146, 220, 247}
\definecolor{mycyan}{RGB}{151, 251, 241}
\definecolor{mygreen}{RGB}{220, 254, 141}
\definecolor{mygreen2}{RGB}{237, 254, 198}
\definecolor{myblue4}{RGB}{196, 225, 255}
\definecolor{myblue5}{RGB}{36,80,138}
\definecolor{under}{RGB}{116,141,164}
\begin{table*}[ht]
\caption{F1-score ± STD comparison(\%) under the standard setting of transductive node classification task with node unlearning request. The highest results are highlighted in \colorbox{mycyan!40}{bold}, while the second-highest results are marked with \colorbox{myblue!20}{\underline{underline}}.}
\centering

\renewcommand{\arraystretch}{1.2} 
\begin{tabular*}{0.99\textwidth}{lcccccccccc}
\arrayrulecolor{myblue5}\toprule[0.16em]
\textbf{Node-Level} &\textbf{Cora}&\textbf{Citeseer} & \textbf{PubMed}&\textbf{ogbn-arxiv} &\textbf{CS}& \textbf{Flickr}&\textbf{Chameleon} &\textbf{Minesweeper} & \textbf{Tolokers}\\ \arrayrulecolor{under}\midrule[0.1em]

GraphEraser & 81.14{\scriptsize \(\pm\)1.00} & 73.57{\scriptsize \(\pm\)1.25} & 84.68{\scriptsize \(\pm\)0.54} & 62.72{\scriptsize \(\pm\)0.18} & 91.24{\scriptsize \(\pm\)0.08} & 46.93{\scriptsize \(\pm\)0.14} & 45.18{\scriptsize \(\pm\)1.46} & 80.45{\scriptsize \(\pm\)0.08} & 78.36{\scriptsize \(\pm\)0.30}\\
GUIDE       & 73.89{\scriptsize \(\pm\)2.18} & 63.50{\scriptsize \(\pm\)0.71} & 84.02{\scriptsize \(\pm\)0.15} & OOM        & 86.96{\scriptsize \(\pm\)0.15} & OOM & {43.37{\scriptsize \(\pm\)0.04}} & 44.71{\scriptsize \(\pm\)0.00} & 44.11{\scriptsize \(\pm\)0.00}\\
GraphRevoker& 81.09{\scriptsize \(\pm\)1.63} & 73.45{\scriptsize \(\pm\)0.61} & 84.94{\scriptsize \(\pm\)0.14} & 62.72{\scriptsize \(\pm\)0.18} & 91.26{\scriptsize \(\pm\)0.10} & 46.91{\scriptsize \(\pm\)0.14} & 44.87{\scriptsize \(\pm\)1.72} & 80.44{\scriptsize \(\pm\)0.12} & 78.36{\scriptsize \(\pm\)0.30}\\ \arrayrulecolor{myblue4}\midrule[0.08em] \addlinespace[-1.5pt]
\arrayrulecolor{under} \midrule[0.08em]
GIF         & 81.75{\scriptsize \(\pm\)1.09} & 62.58{\scriptsize \(\pm\)0.67} & 78.60{\scriptsize \(\pm\)0.22} & 65.52{\scriptsize \(\pm\)0.17} & 91.87{\scriptsize \(\pm\)0.22} & 47.56{\scriptsize \(\pm\)0.12} & 55.09{\scriptsize \(\pm\)1.23}& 77.83{\scriptsize \(\pm\)0.09} & 78.44{\scriptsize \(\pm\)0.10}\\
CGU         & 86.37{\scriptsize \(\pm\)0.78} & \cellcolor{myblue!20}\underline{75.62\scriptsize \(\pm\)0.41} & 76.07{\scriptsize \(\pm\)0.15} & OOT & OOT & OOT & {35.83{\scriptsize \(\pm\)0.74}} & {80.85{\scriptsize \(\pm\)0.00}} & {78.91{\scriptsize \(\pm\)1.49}}\\
ScaleGUN    & 78.82{\scriptsize \(\pm\)0.14} & {73.42\scriptsize \(\pm\)0.13} & 77.88{\scriptsize \(\pm\)0.02} &43.01 {\scriptsize \(\pm\)0.01} & 91.44{\scriptsize \(\pm\)0.09} & 32.95{\scriptsize \(\pm\)0.03} & {49.61{\scriptsize \(\pm\)0.58}} &{44.10{\scriptsize \(\pm\)0.00}} & {59.31{\scriptsize \(\pm\)0.00}} \\
IDEA        & 87.71{\scriptsize \(\pm\)0.25} & 63.66{\scriptsize \(\pm\)0.49} & 80.44{\scriptsize \(\pm\)0.13} & 64.22{\scriptsize \(\pm\)0.17} & 89.47{\scriptsize \(\pm\)0.22} & 42.01{\scriptsize \(\pm\)0.04} & 52.89{\scriptsize \(\pm\)0.80} & 77.93{\scriptsize \(\pm\)0.04} & 78.72{\scriptsize \(\pm\)0.15}\\ 
CEU        & 87.12{\scriptsize \(\pm\)0.07} & 71.56{\scriptsize \(\pm\)0.15} & \cellcolor{mycyan!40}\scalebox{0.95}{\textbf{86.91{\scriptsize \(\pm\)0.06}}} & 57.80{\scriptsize \(\pm\)0.83} & 92.23{\scriptsize \(\pm\)0.07} & \cellcolor{mycyan!40}\scalebox{0.95}{\textbf{49.47{\scriptsize \(\pm\)0.19}}} & \cellcolor{mycyan!40}\scalebox{0.95}{\textbf{61.89{\scriptsize \(\pm\)0.54}}} & 81.05{\scriptsize \(\pm\)0.04} &78.72{\scriptsize \(\pm\)0.03}\\ \arrayrulecolor{myblue4}\midrule[0.08em] \addlinespace[-1.5pt]
\arrayrulecolor{under} \midrule[0.08em]

GNNDelete   & 74.78{\scriptsize \(\pm\)5.49} & 64.26{\scriptsize \(\pm\)3.82} & 84.82{\scriptsize \(\pm\)1.36} & OOM & 76.26{\scriptsize \(\pm\)2.73} & 42.03{\scriptsize \(\pm\)0.00} & 54.43{\scriptsize \(\pm\)1.87} & 81.04{\scriptsize \(\pm\)0.19} & 79.12{\scriptsize \(\pm\)0.16}\\
MEGU        & 82.68{\scriptsize \(\pm\)1.56} & 63.60{\scriptsize \(\pm\)1.11} & 79.68{\scriptsize \(\pm\)0.60} & \cellcolor{myblue!20}\underline{66.15{\scriptsize \(\pm\)0.29}} & 91.69{\scriptsize \(\pm\)0.08} & 47.97{\scriptsize \(\pm\)0.13} & 53.82{\scriptsize \(\pm\)1.68} & 77.64{\scriptsize \(\pm\)0.02} & \cellcolor{mycyan!40}\scalebox{0.95}{\textbf{79.29{\scriptsize \(\pm\)0.10}}}\\
SGU         & \cellcolor{mycyan!40}\scalebox{0.95}{\textbf{89.26{\scriptsize \(\pm\)0.49}}} & 72.04{\scriptsize \(\pm\)1.70} & \cellcolor{myblue!20}\underline{86.61{\scriptsize \(\pm\)0.02}} & \cellcolor{mycyan!40}\scalebox{0.95}{\textbf{67.20{\scriptsize \(\pm\)0.08}}} & \cellcolor{mycyan!40}\scalebox{0.95}{\textbf{93.20{\scriptsize \(\pm\)0.07}}} & \cellcolor{myblue!20}\underline{48.70{\scriptsize \(\pm\)0.21}} & \cellcolor{myblue!20}\underline{60.18{\scriptsize \(\pm\)0.26}} & \cellcolor{myblue!20}\underline{81.11{\scriptsize \(\pm\)0.02}} & \cellcolor{myblue!20}\underline{79.18{\scriptsize \(\pm\)0.04}}\\
D2DGN       & \cellcolor{myblue!20}\underline{88.41{\scriptsize \(\pm\)0.20}} & {73.81{\scriptsize \(\pm\)0.28}} & 86.06{\scriptsize \(\pm\)0.05} & 66.08{\scriptsize \(\pm\)0.18} & \cellcolor{myblue!20}\underline{92.99{\scriptsize \(\pm\)0.03}} & 48.01{\scriptsize \(\pm\)0.03} & 53.25{\scriptsize \(\pm\)0.70} & \cellcolor{mycyan!40}\scalebox{0.95}{\textbf{81.19{\scriptsize \(\pm\)0.04}}} & 79.18{\scriptsize \(\pm\)0.05}\\
GUKD        & 79.65{\scriptsize \(\pm\)1.98} & 70.63{\scriptsize \(\pm\)0.93} & 83.37{\scriptsize \(\pm\)0.60} & OOM & {75.04{\scriptsize \(\pm\)0.82}} & 42.03{\scriptsize \(\pm\)0.04} & {36.62{\scriptsize \(\pm\)2.12}} & 80.89{\scriptsize \(\pm\)0.04} & 78.91{\scriptsize \(\pm\)0.00}\\ \arrayrulecolor{myblue4}\midrule[0.08em] \addlinespace[-1.5pt]
\arrayrulecolor{under} \midrule[0.08em]
Projector    & 86.79{\scriptsize \(\pm\)2.37} & \cellcolor{mycyan!40}\scalebox{0.95}{\textbf{77.00{\scriptsize \(\pm\)0.65}}} & 83.97{\scriptsize \(\pm\)0.30} & OOT & 88.40{\scriptsize \(\pm\)0.63} & OOT & 43.29{\scriptsize \(\pm\)1.17} & 80.85{\scriptsize \(\pm\)0.00} & 78.91{\scriptsize \(\pm\)0.00}\\
\arrayrulecolor{myblue5}\bottomrule[0.16em]
\end{tabular*}
\label{node_node}
\end{table*}

 \textbf{Effectiveness.}
For the effectiveness of algorithms within OpenGU, we conduct evaluations tailored to key downstream tasks while specifically examining GU performance on retained data. We leverage F1-score, AUC-ROC, and Accuracy to evaluate the model's predictive performance at the node, edge, and graph levels, respectively. To evaluate unlearning effects more rigorously, we incorporate membership inference attack (MIA) and poisoning attack (PA). MIA examines whether specific nodes are in the training set, with an AUC-ROC close to 0.5 indicating effective unlearning. Poisoning attack, on the other hand, degrades model predictions by introducing mismatched edges. Improved link prediction after removing these edges validates the algorithm’s ability to erase unwanted relationships effectively. This multi-faceted approach provides a thorough assessment of GU effectiveness, ensuring comprehensive evaluation of both retained and unlearned information. Detailed explanation of metrics and attacks is provided in \cite{OpenGU2025} (A.4 and A.5).

 \textbf{Efficiency.} 
In evaluating the efficiency of GU algorithms in OpenGU, we focus on scalability, time complexity, and space complexity. Scalability assesses each method’s adaptability to different dataset sizes, offering insight into performance stability across varying graph scales. Time complexity analysis includes both theoretical and empirical evaluation to understand computational demands. For space complexity, we examine memory efficiency by measuring peak memory usage and storage requirements during unlearning, determining which algorithms are viable in resource-limited environments. Together, these metrics provide a comprehensive view of each method’s suitability for real-time and scalable deployment.

 \textbf{Robustness.}
To evaluate the robustness of GU algorithms in OpenGU, we systematically examine model performance under varying levels of deletion intensity, noise and sparsity. This involves assessing how different proportions of data perturbation affect the model’s predictive capabilities. Robust GU algorithms should ideally demonstrate minimal performance degradation as deletion intensity increases, reflecting strong resilience in maintaining effective predictions for retained entities.

\section{Experiments and Analyses}
\definecolor{myblue}{RGB}{146, 220, 247}
\definecolor{mycyan}{RGB}{151, 251, 241}
\definecolor{mygreen}{RGB}{220, 254, 141}
\definecolor{mygreen2}{RGB}{237, 254, 198}
\definecolor{myyellow}{RGB}{249, 208, 0}
\definecolor{myyellow2}{RGB}{254, 245, 201}
\definecolor{myyellow3}{RGB}{252, 238, 149}
\definecolor{myblue2}{RGB}{196, 225, 255}

In this section, we delve into a series of targeted experiments designed to rigorously evaluate the effectiveness, efficiency, and robustness of GU algorithms within OpenGU. By posing key questions, we aim to uncover insights into how these algorithms respond to diverse unlearning scenarios, data complexities, and practical deployment challenges, ultimately providing a comprehensive understanding of their efficacy. More detailed information about the experimental setting is presented in \cite{OpenGU2025} (A.6).

For \textbf{effectiveness}, \textbf{Q1:} How effective are GU algorithms in predicting retained data under different unlearning requests? \textbf{Q2:} Do existing GU strategies achieve forgetting in response to unlearning requests? \textbf{Q3:} Do current GU algorithms effectively balance the trade-off between forgetting and reasoning? For \textbf{efficiency},\textbf{Q4:} How do GU algorithms perform in terms of space and time complexity theoretically? \textbf{Q5:} How do the GU algorithms perform regarding space and time consumption in practical scenarios? For \textbf{robustness}, \textbf{Q6:} As the intensity of forgetting requests increases, can the GU algorithms still maintain their original performance levels? \textbf{Q7:} How do GU algorithms perform under sparse and noisy settings?

\subsection{Reasoning Performance Comparison}
To address \textbf{Q1}, we conducted a comprehensive comparison and analysis of existing GU methods across three representative combinations of downstream tasks and unlearning requests.
For clarity, we denote these combinations as downstream task-unlearning request (e.g. node-node). In the subsequent sections, we will conduct an in-depth comparison and analysis of GU algorithms centered around a meticulously designed experimental setup, ultimately deriving insightful conclusions.

Previous studies on GU have often employed varying dataset splits, different GNN backbones, and inconsistent unlearning request configurations, hindering direct comparisons between different methods. To begin with, we outline the unified experimental setup employed in our study. We design tasks across three levels (i,e, node, edge, and graph) using datasets split into 80\% for training and 20\% for testing. For unlearning requests, 10\% of nodes and edges are selected for removal, while in the graph-feature experiments, we randomly select half of the training graphs and set 10\% of their node features to zero vectors. Regarding backbone selection, we leverage SGC as a representative of decoupled GNNs for the node-node task, the classic sampling-based GraphSAGE for the edge-edge task, and the prevalent GCN for the graph-feature task. These tasks correspond to widely used evaluation metrics: F1-score, AUC-ROC, and ACC, respectively. 
For each GU method and dataset, we report the mean performance and standard deviation over 10 runs, ensuring consistency and reliability in the evaluation.

\textbf{Node-Node Experiment.} 
Most existing GU methods are evaluated primarily on node classification tasks. Leveraging the fair comparison environment provided by OpenGU, we evaluate $14$ GU methods across $9$ datasets, with the results presented in Table \ref{node_node}. Our analysis reveals several key observations: (1) The best and second-best results are predominantly achieved by Learning-based methods, with SGU and D2DGN standing out. This indicates that the targeted strategies and loss function designs in Learning-based approaches effectively maintain SOTA performance for predicting retained data.
(2) On smaller, commonly used datasets such as Cora, Citeseer, and PubMed, nearly all methods deliver relatively strong performance. 
However, on larger datasets like ogbn-arxiv, several methods, including CGU, GNNDelete, GUKD, and Projector, encounter challenges such as out-of-memory (OOM) or out-of-time (OOT) issues, with CGU being particularly affected. 

\textbf{Edge-Edge Experiment.}
Upon reviewing previous work in the field of GU, it is found that apart from methods specifically designed for edge unlearning requests, such as GNNDelete and UtU, few existing GU methods have been evaluated on link prediction as a downstream task. This limitation prevents a fair assessment of whether these methods are suitable for link prediction tasks and edge unlearning requests. To address this gap, OpenGU extends the existing collection of GU methods by adapting them for edge-edge unlearning requests, offering a unified interface for seamless implementation of edge-edge experiments.

We conducted experiments on $12$ GU methods across $6$ datasets. Based on the results presented in Table \ref{edge_edge}, several key observations can be drawn: (1) IF-based methods demonstrate exceptional performance in edge-edge experiments, maintaining high accuracy in link prediction after unlearning. Notably, GIF and IDEA stand out, indicating that IF-based approaches exhibit strong generalizability to edge-edge tasks by leveraging influence functions. (2) Partition-based methods face challenges in effectively addressing link prediction tasks. This is primarily due to the inherent sparsity of edges in most datasets, where partitioning further reduces the number of meaningful edges, resulting in an even more limited set of samples. 
(3) While Learning-based methods excel in node-node scenarios, most of them fall significantly behind SOTA in edge-edge settings. However, edge-specific methods like GNNDelete consistently maintain high accuracy across all tested datasets. 

\definecolor{myblue4}{RGB}{196, 225, 255}
\definecolor{myblue5}{RGB}{36,80,138}
\definecolor{under}{RGB}{116,141,164}
\begin{table}[]
\centering
\caption{AUC-ROC ± STD comparison(\%) under the setting of inductive link prediction task with edge unlearning. }
\renewcommand{\arraystretch}{1.2}
\resizebox{\linewidth}{!}{%
\begin{tabular}{lcccccc}
\arrayrulecolor{myblue5}\toprule[0.16em]
\textbf{Edge-Level }  &\textbf{ Cora} & \textbf{CiteSeer} & \textbf{PubMed} & \textbf{DBLP} & \textbf{Physics} & \textbf{Questions} \\ \arrayrulecolor{under}\midrule[0.1em]
GraphEraser  & 67.19{\scriptsize \(\pm\)1.59} & 67.89{\scriptsize \(\pm\)2.60} & 78.53{\scriptsize \(\pm\)0.28} & 74.54{\scriptsize \(\pm\)0.32} & 78.54{\scriptsize \(\pm\)0.69} & 70.16{\scriptsize \(\pm\)0.73}\\
GUIDE        & 64.11{\scriptsize \(\pm\)1.19} & 65.31{\scriptsize \(\pm\)1.99} & 64.67{\scriptsize \(\pm\)0.28} & 65.28{\scriptsize \(\pm\)0.36} & 50.00{\scriptsize \(\pm\)0.08} & 39.32{\scriptsize \(\pm\)0.43}  \\
GraphRevoker & 76.39{\scriptsize \(\pm\)1.18} & 76.00{\scriptsize \(\pm\)1.60} & 87.73{\scriptsize \(\pm\)0.39} &81.27{\scriptsize \(\pm\)0.61} & OOM & OOM   \\ \arrayrulecolor{myblue4}\midrule[0.08em] \addlinespace[-1.5pt] 
\arrayrulecolor{under} \midrule[0.08em]
GIF & \cellcolor{mycyan!40}\scalebox{0.95}{\textbf{84.58{\scriptsize \(\pm\)13.35}}} & \cellcolor{mycyan!40}\scalebox{0.95}{\textbf{85.91{\scriptsize \(\pm\)1.21}}} & \cellcolor{myblue!20}\underline{88.48{\scriptsize \(\pm\)0.95}} & 89.91{\scriptsize \(\pm\)1.49} & \cellcolor{myblue!20}\underline{88.43{\scriptsize \(\pm\)2.28}} & 85.71{\scriptsize \(\pm\)0.21}  \\
ScaleGUN & 78.78{\scriptsize \(\pm\)0.12} & 74.62{\scriptsize \(\pm\)0.31} & 77.89{\scriptsize \(\pm\)0.02} & 73.21{\scriptsize \(\pm\)0.04} & \cellcolor{mycyan!40}\scalebox{0.95}{\textbf{91.27{\scriptsize \(\pm\)0.04}}} & 71.58{\scriptsize \(\pm\)0.09} \\
IDEA & \cellcolor{myblue!20}\underline{84.34{\scriptsize \(\pm\)1.41} }& \cellcolor{myblue!20}\underline{85.41{\scriptsize \(\pm\)0.45}} & \cellcolor{mycyan!40}\scalebox{0.95}{\textbf{88.63{\scriptsize \(\pm\)1.07}}} & \cellcolor{mycyan!40}\scalebox{0.95}{\textbf{92.64{\scriptsize \(\pm\)0.10}}} & 83.96{\scriptsize \(\pm\)0.15} & \cellcolor{myblue!20}\underline{{85.75{\scriptsize \(\pm\)0.20}}}
\\ \arrayrulecolor{myblue4}\midrule[0.08em] \addlinespace[-1.5pt] 
\arrayrulecolor{under} \midrule[0.08em]
GNNDelete & 84.28{\scriptsize \(\pm\)1.38} & 83.47{\scriptsize \(\pm\)2.20} & 86.33{\scriptsize \(\pm\)0.79} & \cellcolor{myblue!20}\underline{90.62{\scriptsize \(\pm\)0.95}} & 83.47{\scriptsize \(\pm\)0.44} &83.52{\scriptsize \(\pm\)1.09}
\\
MEGU & 62.42{\scriptsize \(\pm\)0.86} & 69.83{\scriptsize \(\pm\)0.35} & 57.79{\scriptsize \(\pm\)0.77} & 51.60{\scriptsize \(\pm\)0.23} & 51.21{\scriptsize \(\pm\)0.14} & \cellcolor{mycyan!40}\scalebox{0.95}{\textbf{85.79{\scriptsize \(\pm\)0.25}}}\\
SGU     &78.85{\scriptsize \(\pm\)1.08}     &      78.85{\scriptsize \(\pm\)0.96}&    77.02{\scriptsize \(\pm\)0.23}       & 74.55{\scriptsize \(\pm\)0.38}        &72.66{\scriptsize \(\pm\)0.13}       &  66.68{\scriptsize \(\pm\)0.11}               \\
D2DGN & 80.65{\scriptsize \(\pm\)0.82} & 82.64{\scriptsize \(\pm\)0.92} & 78.90{\scriptsize \(\pm\)2.57} & 83.92{\scriptsize \(\pm\)0.83} & 82.82{\scriptsize \(\pm\)0.43} & 75.53{\scriptsize \(\pm\)0.34}
 \\
GUKD & 56.43{\scriptsize \(\pm\)1.21} & 62.21{\scriptsize \(\pm\)7.09} & 62.21{\scriptsize \(\pm\)7.09} & 78.49{\scriptsize \(\pm\)1.91} & 76.52{\scriptsize \(\pm\)4.79} & 78.83{\scriptsize \(\pm\)0.98}\\
\arrayrulecolor{myblue4}\midrule[0.08em] \addlinespace[-1.5pt] 
\arrayrulecolor{under} \midrule[0.08em]
UtU & 80.76{\scriptsize \(\pm\)1.72} & 82.60{\scriptsize \(\pm\)1.29} & 86.99{\scriptsize \(\pm\)0.88} & 85.95{\scriptsize \(\pm\)0.18} & 72.62{\scriptsize \(\pm\)0.13} & 79.50{\scriptsize \(\pm\)0.67}\\
\arrayrulecolor{myblue5}\bottomrule[0.16em]
\end{tabular}
}

\label{edge_edge}
\end{table}

\begin{figure}[bp]
    \centering
    \includegraphics[width=0.5\textwidth]{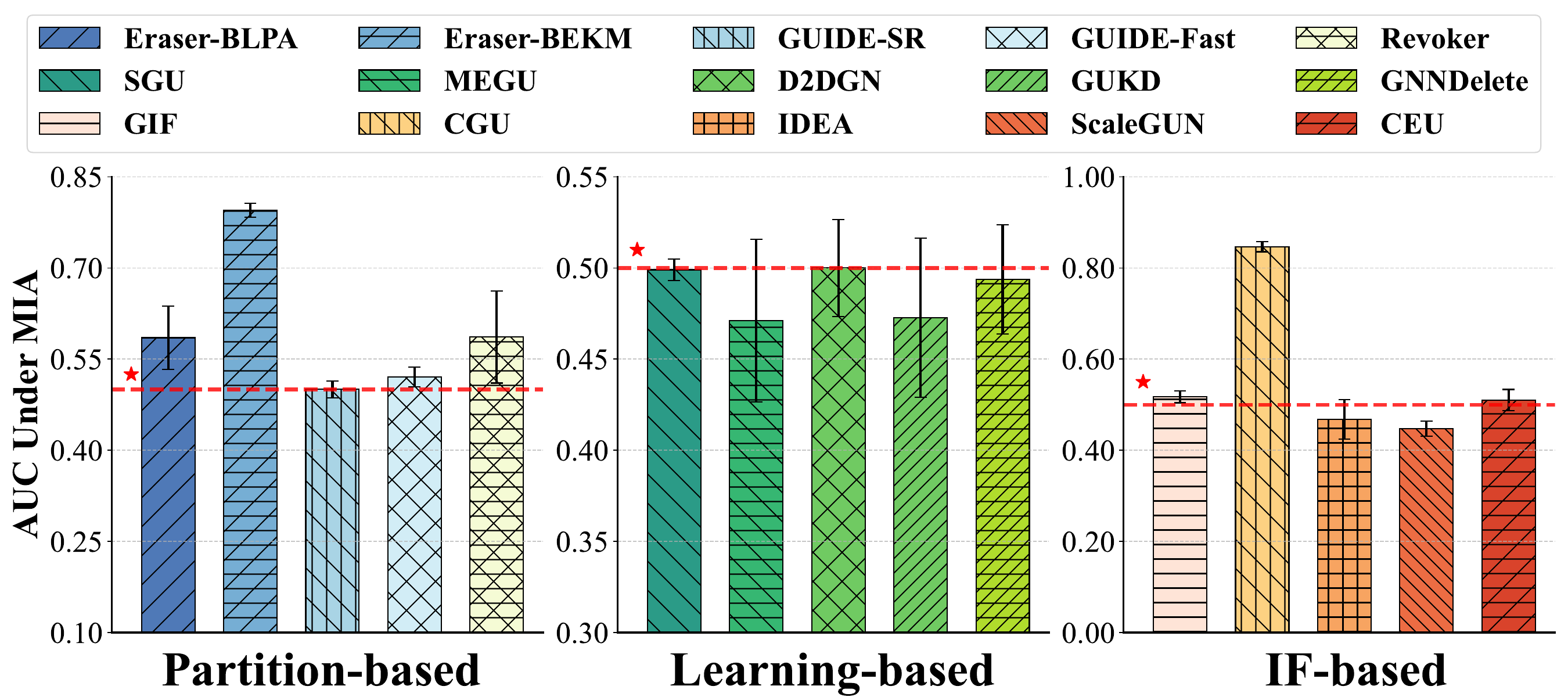}  
    \caption{AUC-ROC ± STD comparison under MIA for node-node task with SGC backbone.}
    \label{MIA}
    \vspace{-0.3cm}
\end{figure}
\textbf{Graph-Feature Experiment.}
In the current landscape of GU research and OpenGU's integration, GST is the only method proposed to address the GU challenge in graph classification task, highlighting a significant area for further exploration. Given the limited availability of methods tailored to such problems, OpenGU extends existing frameworks and GU strategies to enable graph unlearning in graph classification task, implementing four GU methods. Among these, Partition-based approaches are adapted by extending GraphEraser to the graph classification context, where K-means clustering is used to partition graphs for unlearning. GIF and IDEA, on the other hand, focus on leveraging influence functions to update parameters at the graph level. However, learning-based methods were not adapted as their unlearning strategies are not suitable for graph classification tasks (e.g. lack of node class label). 

We evaluate the performance of these $4$ GU methods on $7$ graph datasets, with the results presented in Table \ref{graph_feature}. The findings reveal that IF-based methods remain effective for graph classification task, achieving commendable performance in most cases. While the Partition-based approach is relatively straightforward, GraphEraser shows competitive performance across these datasets. 
However, GraphEraser faces challenges when applied to large-scale graph, as evidenced by the OOM issue observed on the NCI1 dataset.

Through our exploration of \textbf{Q1}, we derive two key conclusions. \textbf{C1}: \textit{Partition and IF-based methods demonstrate broad applicability to most tasks, while Learning-based methods, Projector, and UtU are more specialized, limiting their scalability and generalization} \cite{generalization2,generalization3}. \textbf{C2}: \textit{Learning-based methods excel in targeted tasks, often achieving SOTA performance, while IF-based methods offer flexibility and maintain competitiveness in more tasks} \cite{generalization1}.

\begin{table}[]
\centering
\caption{ACC ± STD comparison(\%) under the setting of graph classification task with feature unlearning. }
\label{graph_feature}
\resizebox{\linewidth}{!}{%
\begin{tabular}{lcccc}
\arrayrulecolor{myblue5}\toprule[0.16em]
\textbf{Graph-Level} & \textbf{GraphEraser}   & \textbf{GST }        &\textbf{ GIF}                     & \textbf{IDEA   }                                \\ \arrayrulecolor{under}\midrule[0.1em]
MUTAG     & 65.79{\scriptsize \(\pm\)0.05}          & 68.42{\scriptsize \(\pm\)0.13}         & 73.68{\scriptsize \(\pm\)1.03}         & \cellcolor{mycyan!40}\scalebox{0.95}{\textbf{73.68{\scriptsize \(\pm\)0.03}}}         \\
PTC-MR    & 53.04{\scriptsize \(\pm\)0.34}          & 55.07{\scriptsize \(\pm\)0.76}         & 60.29{\scriptsize \(\pm\)2.17}         & \cellcolor{mycyan!40}\scalebox{0.95}{\textbf{60.58{\scriptsize \(\pm\)2.13} }}        \\
BZR       & 73.33{\scriptsize \(\pm\)0.19}          & 74.07{\scriptsize \(\pm\)0.51}         & \cellcolor{mycyan!40}\scalebox{0.95}{\textbf{75.31{\scriptsize \(\pm\)3.02} }}        & 73.33{\scriptsize \(\pm\)4.18}         \\
COX2      & \cellcolor{mycyan!40}\scalebox{0.95}{\textbf{90.21{\scriptsize \(\pm\)0.76} }}         & 88.30{\scriptsize \(\pm\)0.46}         & 88.51{\scriptsize \(\pm\)1.24}         & 87.87{\scriptsize \(\pm\)1.59}         \\
DHRF      & 85.26{\scriptsize \(\pm\)2.68}          & \cellcolor{mycyan!40}\scalebox{0.95}{\textbf{86.18{\scriptsize \(\pm\)3.14} }}        & 74.47{\scriptsize \(\pm\)5.32}         & 67.24{\scriptsize \(\pm\)8.28}         \\
AIDS      & 93.30{\scriptsize \(\pm\)0.10}          & 90.90{\scriptsize \(\pm\)0.20}         & 98.40{\scriptsize \(\pm\)0.12}         & \cellcolor{mycyan!40}\scalebox{0.95}{\textbf{98.50{\scriptsize \(\pm\)0.32} }}        \\
NCI1      & OOM                                    & \cellcolor{mycyan!40}\scalebox{0.95}{\textbf{59.08{\scriptsize \(\pm\)0.20}}}         & 52.90{\scriptsize \(\pm\)1.40}         & 53.63{\scriptsize \(\pm\)1.11}         \\ 
\arrayrulecolor{myblue5}\bottomrule[0.16em]
\end{tabular}}

\end{table}

\begin{figure}[bp]
    \centering
    \includegraphics[width=0.5\textwidth]{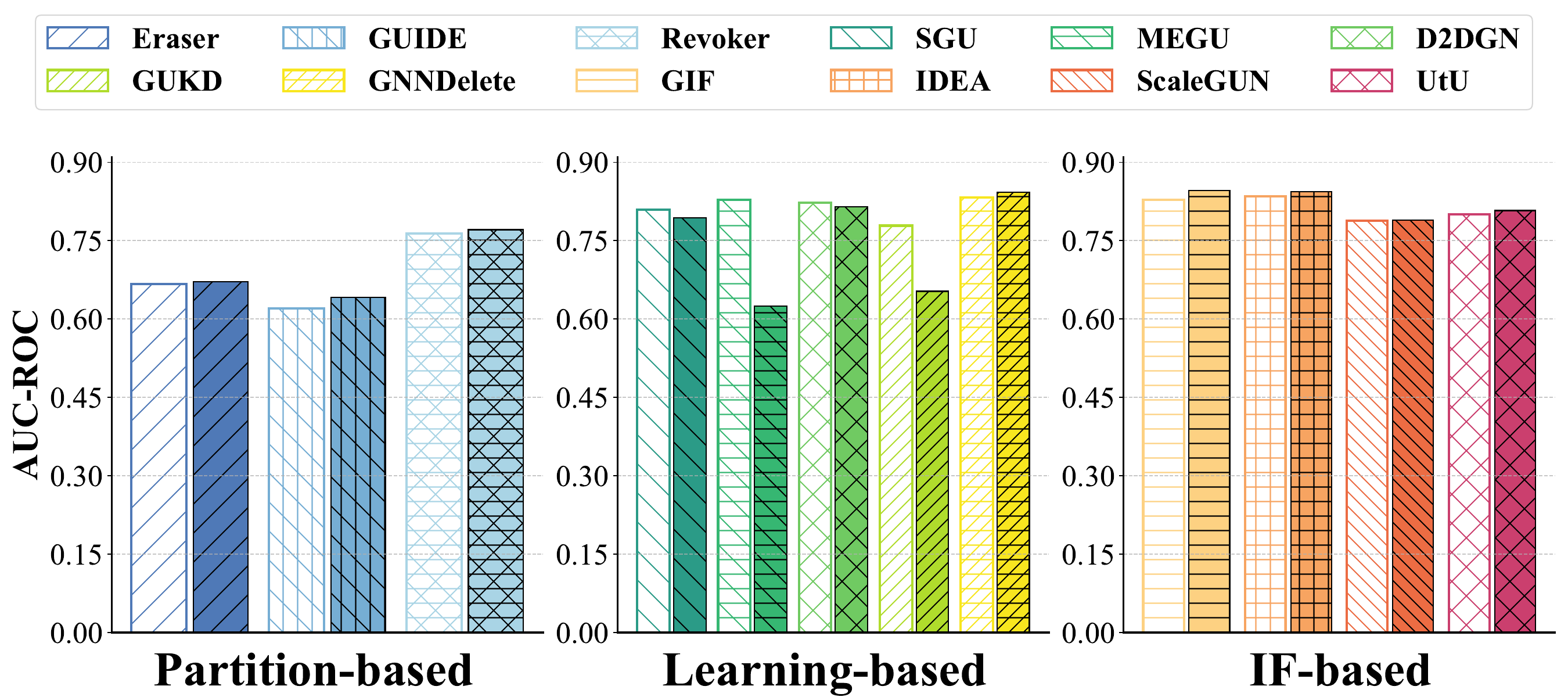}  
    \caption{AUC-ROC comparison under PA for edge-edge task before and after unlearning with GraphSAGE backbone.}
    \label{Poison}
    \vspace{-0.3cm}
\end{figure}

\subsection{Forgetting Performance Comparison}

To address \textbf{Q2}, we evaluate whether GU algorithms effectively forget the unlearning entity by employing commonly used attack strategies in the GU domain, including Membership Inference Attack and Poisoning Attack. These assessments are conducted from both node and edge perspectives to determine whether existing GU methods can genuinely prevent information leakage and protect privacy. The basic experimental setup is the same as Q1.

\begin{figure*}[ht]
    \centering
    \includegraphics[width=0.95\textwidth]{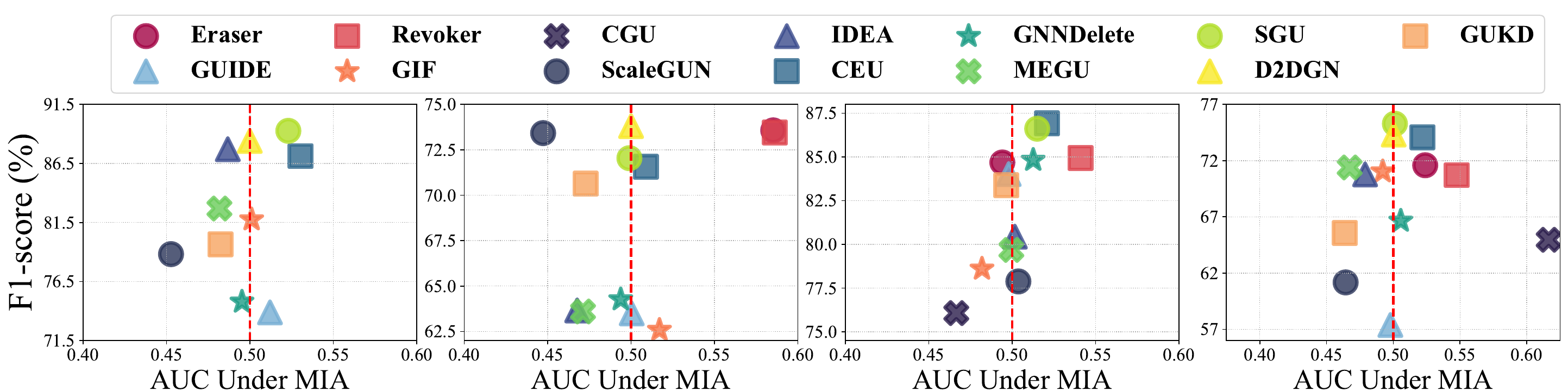}  
    \caption{Trade-off between forgetting and reasoning on Cora, Citeseer, PubMed and in Average performance.}
    \label{Q3}
\end{figure*}

In the node-node experiments, MIA is utilized to assess the extent to which GU algorithms protect model privacy. When the AUC metric approaches 0.5, the predictions become nearly indistinguishable from random guesses, signifying minimal information leakage and thorough removal of sensitive data. To evaluate the performance of various GU methods, we conducted extensive evaluations across multiple datasets and selected the Citeseer dataset as a representative case for result presentation, as shown in Figure \ref{MIA}. The results reveal that: For Partition-based methods, GraphEraser underperforms with the BEKM partitioning strategy, while GUIDE exhibits strong privacy protection when employing SR and Fast partitioning strategies. Among Learning-based methods, SGU, D2DGN, and GNNDelete achieve AUC values close to 0.5, indicating their effectiveness in addressing unlearning requests. Although IF-based methods theoretically analyze and constrain the differences between the original and unlearned models, CGU still falls short of the benchmark for complete unlearning, as evidenced by MIA. 


In the edge-edge experiments, we utilized poisoning attack to evaluate the performance of GU methods. Specifically, we selected nodes from different classes and introduced 10\% heterophilic edges, designated as poisoned edges. Intuitively, the addition of these edges disrupts the message-passing mechanism of GNNs, leading to degraded performance on the link prediction task. These poisoned edges are treated as anomalous data and targeted for removal during unlearning. The effectiveness of unlearning is assessed based on the changes in link prediction AUC; a more significant improvement in prediction accuracy after unlearning indicates that the poisoned edge information has been effectively removed.

Extensive experiments are conducted across various datasets. Observations reveal that smaller datasets are more susceptible to the effects of poisoned edges, resulting in more pronounced performance changes. Consequently, we select the Cora dataset, a relatively small dataset, for the presentation of results, as shown in Table \ref{Poison}. For convenience, UtU is presented in IF-based. The findings show that all Partition-based and IF-based methods achieve an increase in AUC after unlearning, demonstrating their capability to address edge-level unlearning requests in link prediction tasks. Among these, GIF and IDEA stand out by not only effectively removing the harmful edges but also achieving high AUC scores, underscoring their robustness. On the other hand, among the Learning-based methods, only GNNDelete successfully handles the removal of poisoned edges during unlearning, achieving a notably high AUC. Other Learning-based approaches exhibit varying levels of performance decline, suggesting that the unlearning process might inadvertently affect critical information. This highlights the significant challenges Learning-based methods face in effectively addressing such scenarios while maintaining model integrity.

Based on our analysis of \textbf{Q2}, we draw the conclusion \textbf{C3}: \textit{Effectiveness of privacy protection in GU methods is more dependent on the strategy design than on the relational categories. There is still significant potential for improvement in maintaining strong unlearning performance across various scenarios} \cite{privacy1,privacy2}.

\vspace{-0.5cm}

\subsection{Trade-off between Forgetting and Reasoning}
To address \textbf{Q3}, we synthesize the insights from the previous two questions and adopt a unified perspective to analyze their interplay.
Due to space constraints, we focus on the node classification where most GU methods demonstrate strong performance, and utilize SGC as the backbone for consistency across comparisons. To provide a more comprehensive and precise evaluation of the progress achieved by these methods, present the performance of the GU algorithms on Cora, Citeseer, and PubMed respectively. Additionally, we calculate the average performance of the algorithm across nine datasets (consistent with Q1). The summarized results are illustrated in Figure \ref{Q3}, offering a clear overview of the relative performance of different methods in balancing forgetting and reasoning. The methods that are not shown in the figure indicate their AUC exceeds the right boundary.

Given that the most reasonable result of MIA is 0.5, GU methods which are positioned closer to the red centerline and higher up on the graph exhibit superior overall performance. From a vertical perspective, SGU, D2DGN, and CEU excel in reasoning capabilities, whereas GUIDE performs relatively poorly in this regard. From a horizontal perspective, except for CGU, which demonstrates an AUC surpassing 0.6 under MIA, other methods cluster within the range of 0.45 to 0.55. This also indicates the existence of phenomena such as over-forgetting and under-forgetting. Among these, SGU, D2DGN, GNNDelete, and GUIDE manifest robust forgetting capabilities. Notably, while Projector achieves a reasoning performance of approximately 0.68, its approach of projecting weight space into distinct space inadvertently makes sensitive information more accessible to MIA attacks, resulting in an AUC exceeding 0.9. For this reason, Projector is excluded from the comparative visualization.

Upon exploring \textbf{Q3}, we conclude that \textbf{C4}: \textit{Existing GU algorithms still possess room for enhancement in balancing the trade-off between forgetting and reasoning. The issues of under-forgetting and over-forgetting constitute critical challenges that require further investigation to enhance the overall effectiveness} \cite{tradeoff1,tradeoff2}.

\begin{table*}[t]
\caption{\textbf{Algorithm complexity analysis for existing prevalent GU studies.}}
\renewcommand{\arraystretch}{1} 
\resizebox{0.95\textwidth}{!}{
\begin{tabular}{l|cccccc}
\arrayrulecolor{myblue5}\toprule[0.16em]

 \textbf{Method}  & \textbf{Preprocessing}  & \textbf{Training} & \textbf{Unlearning}& \textbf{Inference} & \textbf{Memory} \\
 \arrayrulecolor{under}\midrule[0.1em]
 GUIDE         & $ O(ktn^2\!+\!kctn)$            & $O(Lfn/k\!+\!Lf^2n)$         & $O(Lk'fn^2\!/k^2\!+\!Lkf^2n/k)$ & $ O(Lfm\!+\!Lf^2n\!+\!L_kfn)$         & $O(n^2\!/k^2\!+ \!Lfn\! +\! kn)$ \\
GraphEraser   & $O(kdtn\!+\!ktn \!\log(kn))$     & $O(Lfn/k \!+ \!Lf^2n)$         & $O(Lk'\!fn^2/k^2 \!+ \!Lkf^2n/k)$ & $O(Lfm\!+\!Lf^2n\!+\!Ln_{s}f/k \!+\! Ln_{s}f^2)$& $O(n^2\!/k^2 \!+\! Lfn\! + \!kn)$ \\
 GraphRevoker  & $O(k(d+c)n)$                         & $O(Lfn/k \!+\! Lf^2n)$         & $O(Lk'\!fn^2\!/k^2 \!+\! Lkf^2n/k)$ & $O(Lfm\!+\!Lf^2n\!+\!kf^2n_s)$& $O(n^2\!/k^2\! +\! Lfn \!+ \!kn)$ \\ \arrayrulecolor{myblue4}\midrule[0.08em] \addlinespace[-1.5pt] 
\arrayrulecolor{under} \midrule[0.08em]
 GIF           & -                                               & $O(Lfm \!+\! Lf^2n)$           & $O(n|\theta|)$              & $O(Lfm\!+\!Lf^2n)$ & $O(|\theta|)$ \\
 CGU           & -                                             & $O(Lfm \!+\! f^2n)$            & $O((Lmf\! +\! f^2n)u)$           & $O(Lfm\!+\!f^2n)$   & $O(m \!+\! f^2\! +\! fn)$ \\
 CEU           & -                                              & $O(Lfm \!+\! Lf^2n)$           & $O(t|\theta| \!+\! u|\theta|)$   & $O(Lfm\!+\!Lf^2n)$   & $O(|\theta|)$ \\
 GST           & -                                                &   $O(\sum_{i=0}^{N} pg_i^2)$                       & $O(((p\!+\!|\theta|)\!\sum_{i=0}^{N}g_i^2\!+\!|\theta|^3)u)$        & $O(\sum_{i=0}^{N} pg_i^2)$  & $O(pfn)$ \\
 IDEA          & -                                                & $O(Lfm\!+\!Lf^2n)$                         & $O(n|\theta|)$          & $O(Lfm\!+\!Lf^2n)$ & $O(|\theta|)$ \\
 ScaleGUN      & -                      & $O(Lfm\!+\!f^2n)$       & $O(L^2d^2u)$                          &  $O(Lfm\!+\!f^2n)$ 
  &$O(m\!+\!f^2\!+\!fn)$\\ \arrayrulecolor{myblue4}\midrule[0.08em] \addlinespace[-1.5pt] 
\arrayrulecolor{under} \midrule[0.08em]
 SGU           &$O(Bf^2 \!+\! Bn_s \!+\! f^2u)$                                             & $O(Lfm \!+ \!f^2n)$            & $O(Lf^2\!+\!Bn_sf\!+\!(c\!+\!f)u)$     & $O(Lfm \!+\! f^2n)$  & $O(Bfn_s \!+\! f^2)$ \\
 MEGU          & $O(Lfm \!+ \!Lf^2n\!+\!d^2u)$                                             & $O(Lfm \!+ \!Lf^2n \!+ \!d^2u)$            &  $O(d^2cu)$     & $O(Lfm \!+\! Lf^2n)$ & $O(Lfm \!+ \!f^2n)$ \\
 GUKD          &   $O(Lfm\!+\!Lf^2n)$                                             &$O(L_t fm\!+\!Lf^2n)$                        & $O(c(n\!-\!u))$                     & $O(Lfm \!+\! Lf^2n)$  & $O(f^2\!+\!fn)$\\
 D2DGN         &       $O(Lfm\!+\!Lf^2n)$                    &  $O(Lfm\!+\!Lf^2n) $                     &              $O(c(n\!-\!u))$         & $O(Lfm \!+\! Lf^2n)$  & $O(f^2\!+\!fn)$ \\
 GNNDelete     & $O(Lfm\! +\! Lf^2n \!+\!d^2u)$                                             & $O(Lfm \!+\! Lf^2n)$             & $O(d^2fu)$ & $O(Lfm \!+\! Lf^2n)$ & $O(fu \!+ \!d^2\!fu)$ \\ \arrayrulecolor{myblue4}\midrule[0.08em] \addlinespace[-1.5pt] 
\arrayrulecolor{under} \midrule[0.08em]
 Projector     & -                                             & $O(Lfm \!+\! Lf^2n)$           & $O(f^2n \!+\! \max\{u^3\!, \!f^2u\})$ & $O(Lfm \!+\! Lf^2n)$  & $O(f^2 \!+ \!fn)$ \\
 UtU           & -                                             & $O(Lfm \!+\! Lf^2n)$                          & $O(u)$                        & $O(Lfm \!+\! Lf^2n)$  & $O(m \!+\! Lfn)$ \\ 
 \arrayrulecolor{myblue5}\bottomrule[0.16em]
\end{tabular}%
}
\label{complex}
\vspace{-0.15cm}
\end{table*}

\vspace{-1.5em}
\subsection{Algorithm Complexity Analyses}
To answer \textbf{Q4} about how the GU algorithms perform in terms of time and space complexity, we analyze the time complexity from the perspectives of preprocessing, training, unlearning, and inference. Since partition-based methods are unique, their partitioning is included in the preprocessing, while aggregation is incorporated into inference.
Additionally, we provide a comprehensive summary of the space complexity for all methods, encompassing the required memory for model parameters, data storage, and auxiliary structures, ensuring a thorough evaluation. Except for the methods that can only act on edge-level, we take node unlearning as the perspective of analysis. The results are presented in Table \ref{complex}.

For clarity in the complexity analysis, we introduce a set of key notations that will be used consistently throughout the discussion. To ensure a uniform evaluation of complexity, we consider GCN as the backbone model, with the number of layers denoted by $L$. In terms of the dataset, $n$ represents the total number of nodes, $m$ refers to the number of edges, $f$ is the feature dimension of the nodes, and $d$ denotes the average degree of the nodes. The number of classes is denoted by $c$, 
while $u$ signifies the number of unlearning requests, representing the nodes, edges or features to be deleted during the unlearning process. 
Since part of the methods involve sampling, we define the number of samples as $n_s$.

For \textbf{Partition-based} methods, the parameter $k$ represents the number of shards, 
and $t$ denotes the number of iterations required by the algorithm. For the three methods considered, the fundamental approach is similar, leading to comparable time and space complexities in both the training and unlearning phases. The principal differences between these GU methods are manifested in the partitioning and aggregation. In the partitioning phase, GUIDE and GraphEraser depend on non-training algorithms, whereas GraphRevoker relies on constructing loss functions. During the aggregation phase, GUIDE utilizes the pyramid match graph kernel, with complexity $ O(L_kfn)$, where $L_k$ is the number of layers in the kernel.

For \textbf{IF-based} methods, the unlearning process relies on influence function, where the time and memory complexity are primarily determined by the model parameters $\theta$. These methods often optimize by approximating the inverse of the Hessian matrix, which typically requires a complexity of $O(|\theta|^3)$. In the case of GST, the initialization time is expressed as $O(\sum_{i=0}^{N} pg_i^2)$, where $N$ is the number of graphs, $g_i$ denotes the node number in graph $G_i$, and $p$ corresponds to $O(\sum_{l=0}^{L-1} J^l)$ determined by the $J$-ary scattering tree.

For \textbf{Learning-based} methods, the training phase typically involves training the GNNs
with the time complexity bounded by $O(Lfm + Lf^2n)$. In the case of GUKD, an additional step is required to train a teacher model, which may differ in layer depth from the target model. To account for this difference, we define the layer count of the teacher model as $L_t$.
The memory complexity of Learning-based GU methods varies depending on their specific design and operations. For example, SGU leverages node influence maximization strategies, resulting in a memory complexity of $O(Bfn_s \!+\! f^2)$, where $B$ stands for the budget hyperparameter. 

Based on the analysis, we conclude \textbf{C5}: \textit{To reduce time costs, partition-based methods require optimization in partitioning, IF-based methods need to minimize the computational overhead of Hessian matrix calculations in specific GU scenarios, and learning-based approaches demand enhanced efficiency in the preprocessing} \cite{Efficient1,Efficient2}.

\begin{figure*}[ht]
    \centering
    \includegraphics[width=0.97\textwidth]{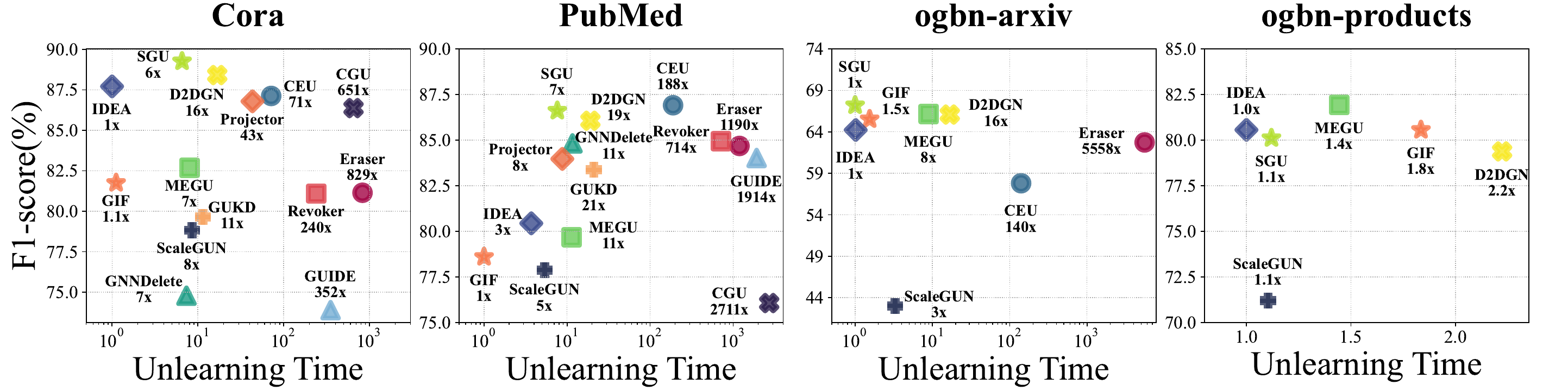}  
    \vspace{-0.1cm}
    \caption{Unlearning Time Performance on Cora, PubMed, ogbn-arxiv and ogbn-products.}
    \label{Q5_time}
    \vspace{-0.1cm}
\end{figure*}
    \vspace{-0.1cm}
\begin{figure*}[ht]
    \centering
    \includegraphics[width=0.97\textwidth]{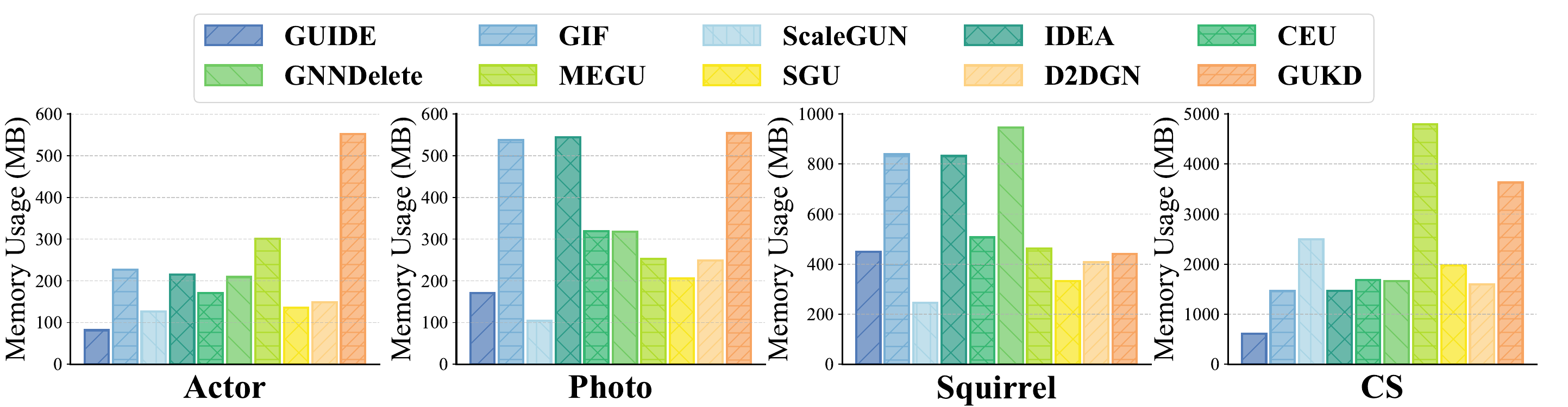}  
    \vspace{-0.1cm}
    \caption{Memory Usage Performance on Various Datasets.}
    \label{Q5_mem}
    \vspace{-0.1cm}
\end{figure*}

\subsection{Practical Efficiency Analyses}
In addressing \textbf{Q5}, we evaluated the time and memory overhead of GU algorithms under a 10\% node unlearning request on a range of datasets, with a focus on the node classification. As depicted in Figure \ref{Q5_time}, we selected four datasets of increasing scale, from the small Cora dataset with thousands of nodes to the large-scale ogbn-products dataset with millions of nodes, to ensure a fair comparison in terms of time cost and scalability. Partition-based methods display substantial time overhead on account of partitioning, aggregation, and shard training operations, whereas most IF-based and Learning-based methods exhibit greater temporal efficiency. On large-scale datasets like ogbn-products, only $6$ GU methods were able to run, with comparable performance, while others faced challenges such as timeouts or memory overflows. Regarding memory usage in Figure \ref{Q5_mem}, GUIDE consistently exhibits low memory overhead across various datasets, while GraphEraser (without visualization) incurs considerably higher costs, exceeding 4000MB even on smallest dataset Actor. Notably, methods that are capable of handling large datasets, such as ScaleGUN and SGU, maintain stable memory consumption across all datasets, thereby demonstrating their scalability. 

Based on our analysis, we conclude \textbf{C6}: \textit{Existing GU methods require further optimization to effectively reduce time and space overhead, emphasizing the need for more efficient implementations in both computation and memory usage} \cite{Efficient3,Efficient4}.

\subsection{Impact of Unlearning Intensity}
\begin{figure*}[ht]
    \centering
    \includegraphics[width=1\textwidth]{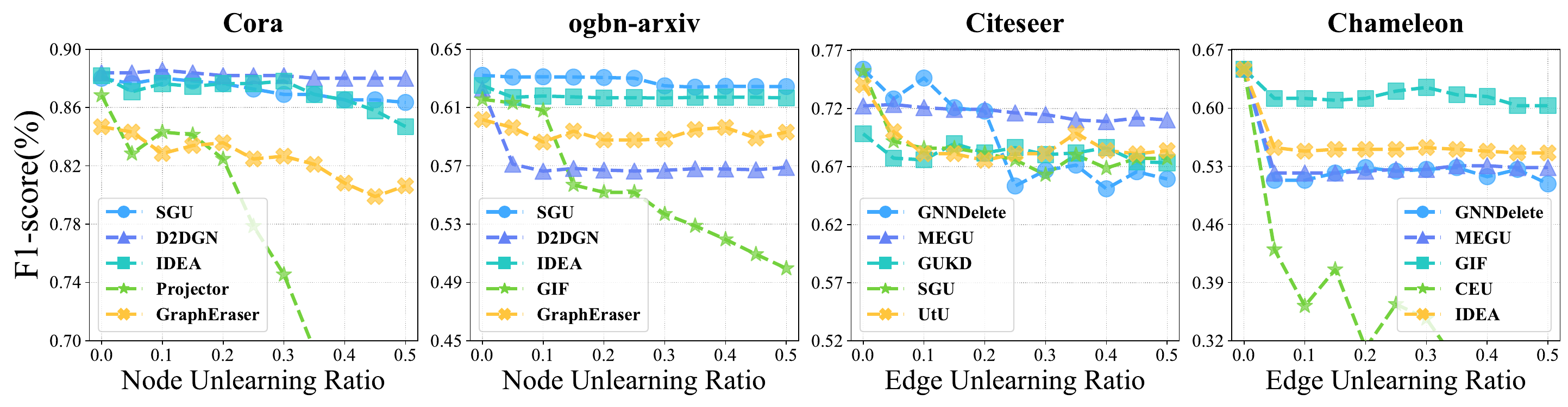}  
    \caption{Performance under Different Unlearning Intensities with GraphSAINT, Cluster-gcn, GAT, and GCN.}
    \label{Q6_ratio}
\end{figure*}

To address Q6, we conducted node unlearning experiments on Cora and ogbn-arxiv and edge unlearning experiments on Citeseer and Chameleon, employing various backbones for the node classification task. The unlearning ratio was incrementally increased from 0 to 0.5, and the results are presented in Figure \ref{Q6_ratio}. The illustration reveals a consistent downward trend in performance for all GU methods as the unlearning ratio increases, indicating that higher deletion intensities negatively impact prediction capabilities. Notably, methods such as Projector, GIF, and CEU exhibit greater sensitivity to changes in certain datasets, while Learning-based methods demonstrate a more gradual decline, highlighting their robustness under higher unlearning intensities. However, we also observe that even with minimal deletion ratios, many methods experience significant performance degradation during unlearning, particularly on the Chameleon dataset. This suggests that current GU algorithms need to take the deletion ratio into account to better reduce the gap in predictive performance between the unlearned and original model. Based on this analysis, we derive conclusion \textbf{C7}: \textit{While most GU algorithms perform reasonably well across varying unlearning intensities, there remains a critical need to enhance their robustness across diverse datasets} \cite{robustness1}.

\begin{figure*}[ht]
    \centering
    \includegraphics[width=1\textwidth]{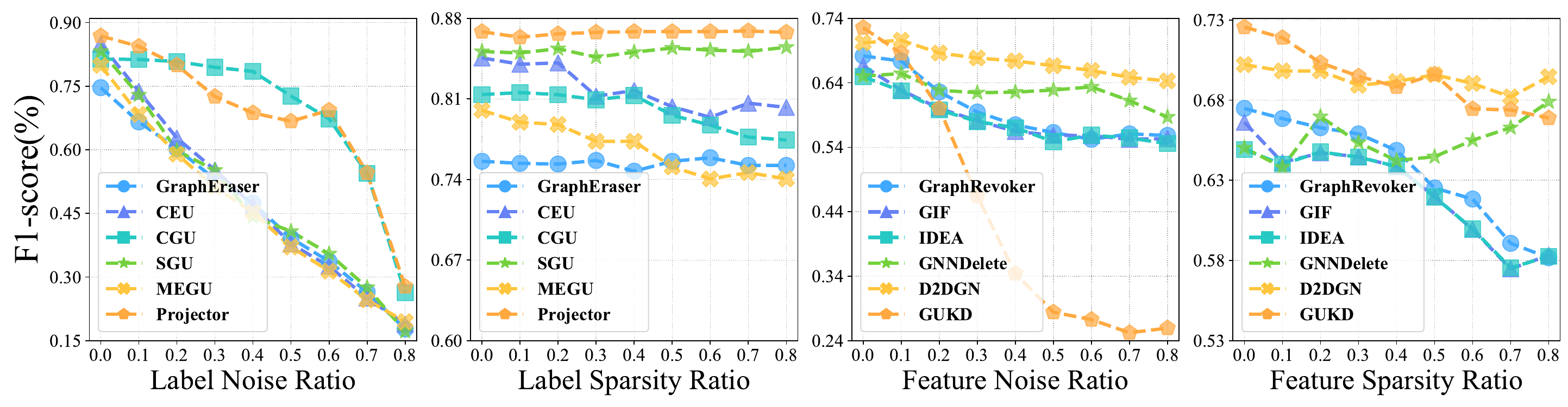}  
    \caption{Performance under Different Noise and Sparsity Ratios at Label and Feature Levels.}
    \label{Q7}
\end{figure*}
\subsection{Robustness Analyses}
To address \textbf{Q7}, we simulate more realistic noise and sparsity scenarios by introducing perturbations at both the label and feature levels to comprehensively evaluate the robustness of existing GU methods. For label noise, a certain proportion of training samples are randomly assigned incorrect labels, while for feature noise, Gaussian noise is injected based on the dimensionality of node features. Sparsity is introduced by varying the proportion of training nodes and simulating partial feature absence. Given the large number of GU methods, we select representative approaches for analysis based on their categories, using the same settings as the node-node experiments in Q1 and adopting SSGC as the backbone. The results, presented in Figure \ref{Q7}, reveal that nearly all methods experience a decline in predictive performance under the influence of noise and sparsity, with label noise exerting the most pronounced impact. Specifically, CGU and Projector exhibit a pattern of slow initial degradation followed by a rapid decline, while other methods demonstrate a steady and sharp drop. When the noise ratio reaches 0.8, the F1-score for all methods falls below 0.3. In contrast, the impact of label sparsity is comparatively minor, with some methods even maintaining their original performance. Under feature noise, GUKD experiences a significant performance drop, indicating its limited resilience to feature perturbations. For feature sparsity, methods like GIF, IDEA, and GraphRevoker show marked declines, whereas GNNDelete surprisingly improves. These findings lead us to conclude \textbf{C8}: \textit{Current GU methods exhibit insufficient robustness to noise and sparsity, particularly in the presence of label noise, which poses a significant challenge to enhancing the overall robustness of GU approaches} \cite{robustness2}.

\vspace{-0.3cm}

\section{Conclusion and Future Directions}
In this paper, we first provide a comprehensive review of the current advancements in the field of Graph Unlearning, highlighting its practical applications and systematically categorizing existing algorithms based on their technical characteristics. Building on this foundation, we introduce OpenGU, the first unified and comprehensive benchmark for GU, which integrates 16 state-of-the-art GU algorithms and 37 datasets across multiple domains. OpenGU further extends the flexibility of GU algorithms, allowing seamless combinations across three downstream tasks and three types of unlearning requests. Using the standardized and fair evaluation framework provided by OpenGU, we conduct extensive experiments to assess the advancements of GU methods from the perspectives of effectiveness, efficiency, and robustness. These experiments not only reveal the notable breakthroughs in the field but also expose critical limitations in existing approaches. To inspire further research, we outline the major challenges currently faced by GU and propose promising directions for future exploration.

\textbf{Designing Generalized GU Frameworks for Diverse Tasks} (\textit{C1} and \textit{C2}). While some existing methods demonstrate strong predictive performance for specific tasks, their underlying principles often make it difficult to adapt to varying downstream tasks and unlearning requests. Furthermore, the current design of unlearning requests remains relatively simplistic, whereas real-world applications often require handling mixed unlearning requests or subgraph-level unlearning. Achieving consistently superior predictive capabilities across such complex and varied scenarios remains a considerable challenge. Therefore, developing more generalized frameworks becomes imperative.

\textbf{Unified Metrics for Evaluating Forgetting} (\textit{C3} and \textit{C4}). The current methods for assessing the forgetting capability of GU approaches remain insufficient and leave significant gaps to be addressed. These methods are often tightly coupled with specific unlearning requests and downstream tasks, making them less effective in handling diverse combinations of scenarios. Moreover, determining whether the information targeted for removal has been thoroughly unlearned from theoretical perspective remains a critical yet underexplored challenge. Furthermore, GU algorithms must strike a balance between reasoning and forgetting. Future research should move beyond the current paradigm of independently assessing these two aspects, striving instead for a unified metric that evaluates models from an integrated perspective, ensuring a comprehensive understanding of their capabilities.

\textbf{Enhancing Algorithm Efficiency} (\textit{C5} and \textit{C6}). While theoretical analysis provides valuable insights, the practical performance of current GU methods often falls short, especially when scaling to large datasets with millions of nodes. These methods commonly encounter OOT or OOM issues. To enable GU's effective deployment in large-scale scenarios, algorithms must be optimized for both efficiency and scalability, ensuring they can handle the demands of real-world data while avoiding these performance bottlenecks.

\textbf{Addressing Realistic Scenarios} (\textit{C7} and \textit{C8}). In practical applications, the presence of noise and incomplete datasets is an unavoidable challenge. However, current GU algorithms lack sufficient exploration and adaptation to such scenarios. Experimental results highlight significant weaknesses when dealing with noise and sparsity, particularly in terms of label and feature robustness. Future research should aim to broaden the scope of investigation, extending robustness analysis to encompass a wider range of real-world challenges and data imperfections.

OpenGU embodies the collective efforts and expertise of the current GU field, offering a comprehensive platform for research and development. As we move forward, we will continue to expand and enhance OpenGU, strengthening its support for future advancements in this area. Finally, we welcome feedback and encourage suggestions to refine our benchmark, further improving its effectiveness and user experience.


\bibliographystyle{ACM-Reference-Format}
\balance
\bibliography{ref}

\clearpage
\appendix
\section{OUTLINE}
\balance
The appendix is organized as follows:

\noindent\textbf{A.1} Dataset Details

\noindent\textbf{A.2} Backbone Details

\noindent\textbf{A.3} GU Method Details

\noindent\textbf{A.4} Evaluation Metric Details

\noindent\textbf{A.5} Attack Details

\noindent\textbf{A.6} Experimental Setting Details

\subsection{Dataset Details}
\label{A.Dataset}
In OpenGU, the selected datasets play a pivotal role in benchmarking and evaluating the performance of GU methods under a range of realistic and challenging conditions. These datasets, chosen for their diversity in structure and application domain, enable a comprehensive analysis of the methods’ effectiveness, efficiency, and robustness. By capturing various graph characteristics, they provide an essential foundation for assessing unlearning strategies and comparing their adaptability across different scenarios. The following section provides a detailed overview of each dataset:

\textbf{Cora}, \textbf{CiteSeer}, and \textbf{PubMed} \cite{Yang16cora} are among the most widely utilized citation network datasets in the GU field. In these datasets, nodes represent research papers, and edges indicate citation relationships between them. Each node is characterized by a bag-of-words feature vector and is uniquely associated with a specific category. These datasets are frequently employed for tasks such as node classification, providing a reliable basis for evaluating model performance across citation networks.

\textbf{DBLP} \cite{2019DBLP}, derived from the extensive DBLP Computer Science Bibliography, offers a unique view into academic collaboration by modeling a co-authorship network. In this dataset, nodes represent individual authors, and edges indicate co-authored publications, capturing dynamic, multi-disciplinary research networks. Each node is labeled according to research domains, enabling robust experiments in node classification, clustering, and link prediction.

\textbf{ogbn-arxiv} \cite{hu2020ogb} is a comprehensive academic graph derived from the Microsoft Academic Graph (MAG) \cite{wang2020microsoft_MAG}, designed to facilitate machine learning tasks on graph data. It represents a paper citation network of arXiv papers, capturing the scholarly communication and influence within the field of computer science. The graph structure of ogbn-arxiv is characterized by nodes representing scientific papers and edges representing citations between them, reflecting the academic referencing relationships. It offers a challenging and realistic testbed for the development and evaluation of GNNs and other machine learning models on graph data.

\textbf{CS} and \textbf{Physics} \cite{shchur2018amazon_datasets} are co-authorship graphs derived from the Microsoft Academic Graph, specifically designed for node classification tasks. These datasets represent academic collaborations where nodes correspond to authors and edges indicate co-authorship on papers. In both datasets, node features are represented by paper keywords associated with each author's publications and class labels denote the most active fields of study for each author, providing a rich semantic profile for classification purposes.

\textbf{Flickr} \cite{zeng2019graphsaint} encapsulates the structure and properties of online images, with each node representing an individual image. The dataset is characterized by its rich feature set, where nodes are described by a comprehensive set of features extracted from image metadata, such as comments, likes, and group affiliations. Edges in the Flickr dataset signify connections between images, which are based on shared attributes or interactions. The Flickr dataset stands as a significant resource for researchers and practitioners in the field of graph neural networks.

\textbf{Photo} and \textbf{Computers} \cite{shchur2018amazon_datasets} datasets are derived from the Amazon co-purchase graph, representing the relationships between products frequently bought together. Nodes in these datasets symbolize individual products, while edges indicate co-purchase frequency, reflecting the market dynamics and consumer behavior on Amazon's e-commerce platform. The Photo dataset focuses on photographic equipment, while the Computers dataset centers on computer-related products, providing a nuanced view into consumer purchasing patterns within these specific domains.

\textbf{ogbn-products} \cite{hu2020ogb} dataset, part of the Open Graph Benchmark (OGB), is an extensive co-purchasing network that captures the intricate relationships between products. Nodes in this dataset symbolize products and edges indicate that two products are frequently bought together. Node features are extracted from product descriptions, transformed into a bag-of-words representation. This dataset is renowned for its large scale and complex structure, which makes it an exemplary testbed for large-scale graph learning applications. It serves as a critical benchmark for assessing the scalability and effectiveness of graph algorithms, providing a realistic challenge for models to handle vast amounts of data and intricate connections.

\textbf{Chameleon} and \textbf{Squirrel} \cite{pei2020geomgcn} datasets, sourced from Wikipedia, represent webpage networks where nodes correspond to individual pages and edges indicate mutual links. Each node’s features are derived from webpage content, capturing unique structural characteristics specific to each network. These datasets are particularly valuable for evaluating heterophilic graph learning methods, as nodes with similar labels are less densely connected, challenging traditional GNNs and encouraging the development of models that effectively handle low homophily settings.

\textbf{Actor} \cite{pei2020geomgcn} dataset represents a actor network within the film industry, where each node corresponds to an actor and edges represent co-appearances in the same movie. Each node is characterized by features based on personal and professional attributes, providing a basis for relational insights. The dataset’s structure, with nodes labeled based on actor roles or genres, is well-suited for evaluating algorithms in low-homophily settings, challenging GNNs to accurately classify nodes when similar labels are sparsely connected.

\textbf{Minesweeper} \cite{platonov2023hete_gnn_survey4} dataset, sourced from an online gaming platform, models interactions within the Minesweeper game environment. In this graph, each node represents a player, and edges denote collaborations or competitions between players during gameplay sessions. Node features are derived from player statistics and gameplay metrics, providing a comprehensive view of user behavior patterns. This dataset is valuable for analyzing patterns in social connectivity and behavior, making it useful for assessing unlearning strategies in dynamic, interaction-driven networks.

\textbf{Tolokers} \cite{platonov2023hete_gnn_survey4} dataset, drawn from a crowdsourcing platform \cite{Tolokers_original}, represents worker interactions within collaborative tasks. Nodes correspond to individual workers, with edges indicating collaborations on shared tasks. The dataset captures complex relationships, as it aims to predict which workers may face restrictions or bans based on past activities.

\textbf{Roman-empire} \cite{platonov2023hete_gnn_survey4} dataset is a dependency graph from the Roman Empire Wikipedia article \cite{lhoest2021empire_original}, with nodes as words and edges as dependencies or word adjacencies. It is used for node classification based on syntactic roles, offering insights into language structure and word relationships within a historical text.

\textbf{Amazon-ratings} \cite{platonov2023hete_gnn_survey4} dataset captures user interactions with products on Amazon, forming a graph where nodes are items and edges represent user ratings. This dataset is instrumental for tasks like predicting user preferences and understanding product affinities, offering a real-world testbed for GNNs to operate at scale.

\textbf{Questions} \cite{platonov2023hete_gnn_survey4} dataset represents interactions within a community Q\&A platform, where nodes correspond to users and edges indicate interactions such as asking, answering, or commenting on questions. This dataset reflects user engagement and information exchange patterns, making it suitable for evaluating algorithms focused on social dynamics and collaborative learning. Its structure provides valuable insights into the spread of information and influence within online communities.

\textbf{MUTAG} \cite{MUTAG} is a chemical compounds dataset focusing on the mutagenicity of molecules, where each graph represents a molecule and the nodes represent atoms. The task is to predict whether the molecule is mutagenic or not. The dataset consists of 188 graphs with 7 distinct classes, making it a benchmark for graph classification in the field of cheminformatics.

\textbf{PTC-MR} \cite{PTC} is a dataset used for mutagenic toxicity prediction, where each graph represents a chemical compound, and the nodes correspond to atoms. The task is to predict whether a compound is mutagenic based on its chemical structure. With 344 graphs, it includes two distinct classes and is commonly used for evaluating GNNs in bioinformatics applications.

\textbf{BZR} \cite{BZR_COX_DHFR} dataset contains chemical compounds, where each molecule is represented as a graph, with nodes representing atoms and edges representing bonds. The classification task is to determine whether a compound can act as a potent estrogen receptor binder. It includes 405 graphs, with active and inactive classes.

\textbf{COX2} \cite{BZR_COX_DHFR} is a dataset related to the inhibition of the COX-2 enzyme, an important target for anti-inflammatory drugs. It consists of chemical compounds where the nodes represent atoms and the edges represent bonds. The task is to predict whether a compound inhibits COX-2 or not. The dataset includes 467 graphs and is used for evaluating graph neural networks in drug discovery.

\textbf{DHFR} \cite{BZR_COX_DHFR} is a dataset of 1,634 chemical compounds used to predict inhibition of the dihydrofolate reductase (DHFR) enzyme, a target for antimicrobial drugs. The dataset is significant for its application in drug design, where GNNs are used to model molecular structures and predict bioactivity. The task is binary classification (inhibitor or non-inhibitor), making it a valuable benchmark in computational biology.

\textbf{AIDS} \cite{AIDS} dataset, which contains 2,000 chemical compounds, is used to predict the activity of these compounds against the HIV virus. It is a key dataset in drug discovery, particularly for anti-HIV drug. Each molecule is represented as a graph where nodes are atoms and edges are bonds. The binary classification task helps evaluate GNNs in pharmaceutical research and HIV treatment.

\textbf{NCI1} \cite{NCI1} dataset consists of 4,110 chemical compounds and is used for cancer cell line screening. Each graph represents a molecule, and the task is to classify whether a compound is active or inactive against cancer cells. It is one of the largest datasets in molecular graph classification, playing a significant role in drug discovery, particularly in identifying potential anticancer compounds.

\textbf{ogbg-molhiv} \cite{hu2020ogb} dataset, part of the Open Graph Benchmark (OGB), consists of 41,127 graphs representing chemical compounds. The goal is to predict whether a compound inhibits HIV. This large-scale dataset is important for advancing GNNs in drug discovery, providing a real-world application in the search for anti-HIV drugs. It is valuable due to its scale and complexity in bioinformatics.

\textbf{ogbg-molpcba} \cite{hu2020ogb} dataset is another graph dataset from the Open Graph Benchmark (OGB) focused on predicting the biological activity of compounds across multiple targets. With over 437,000 compounds, this dataset presents a multi-label classification task, where each compound can have multiple activity labels. It is important for developing models capable of handling large-scale, multi-label classification tasks in cheminformatics.

\textbf{ENZYMES} \cite{chen2024fedgl} dataset is used for enzyme classification, where each graph represents a protein, and the task is to predict the enzyme class to which it belongs. The dataset contains 600 graphs and is significant for understanding protein structures and functions in computational biology. It provides a benchmark for GNNs in biological and biomedical applications, particularly for enzyme function prediction.

\textbf{DD} \cite{DD} dataset consists of 1,178 protein-protein interaction graphs, where nodes represent proteins and edges denote interactions between them. The task is binary classification, determining whether two proteins interact. This dataset is important in the study of biological systems, particularly in understanding cellular processes and identifying potential therapeutic targets.

\textbf{PROTEINS} \cite{PROTEINS} dataset contains 1,113 graphs, each representing a protein, with nodes corresponding to amino acids and edges indicating spatial proximity. The classification task is to distinguish between enzymes and non-enzymes. This dataset is significant for studying protein structure and function, which is fundamental in drug discovery and molecular biology.

\textbf{ogbg-ppa} \cite{hu2020ogb} dataset, from the Open Graph Benchmark (OGB), includes 158,100 graphs representing pairs of proteins and their potential interactions. The task is binary classification to predict whether a pair of proteins interacts. It is particularly valuable for exploring protein interaction prediction, which plays a critical role in understanding biological networks and disease mechanisms.

\textbf{IMDB-BINARY} \cite{IMDB} dataset consists of 1,000 graphs, each representing a movie and the relationships between users and movies. The task is to predict whether a movie belongs to a specific genre based on interactions. This dataset is relevant for applications in social network analysis and movie recommendation systems.

\textbf{IMDB-MULTI} \cite{IMDB} dataset, derived from IMDB, includes 1,500 graphs, where each graph represents a movie and contains co-occurrence relationships between users and movies. The task is to predict multiple genres for each movie. It is important for multi-label classification tasks in social network analysis and content-based recommendation systems.

\textbf{COLLAB} \cite{COLLAB} dataset contains 5,000 graphs, each representing a scientific collaboration network, where nodes represent researchers and edges represent co-authorships. The classification task involves predicting the research domain of a collaboration network. This dataset is valuable for studying academic collaboration patterns and community detection in social networks.

\textbf{ShapeNet} \cite{ShapeNet} dataset is a large-scale collection of 3D models representing various object categories, including chairs, tables, and airplanes. Each object is represented as a graph where nodes correspond to geometric features, and edges define spatial relationships. The task is to classify objects into categories, making it important for 3D shape recognition and computer vision applications.

\textbf{MNISTSuperPixels} \cite{MNISTSuperPixels} dataset is a graph-based version of the famous MNIST digit classification dataset, where each digit is represented as a graph of superpixels. The task is to classify the digits (0-9) based on the graph structure. It is useful for exploring how graph-based models can be applied to image classification tasks, particularly in the context of digit recognition.

\subsection{Backbone Details}
\label{A.Backbone}

In this section, we provide an in-depth overview of the GNNs utilized within OpenGU. These foundational architectures play a critical role in implementing and benchmarking GU strategies, as they offer diverse structures and mechanisms that impact the model's effectiveness, efficiency, and adaptability. By detailing each backbone, we aim to give readers a clearer understanding of the underlying model choices and their relevance to unlearning process.

\textbf{GCN} \cite{kipf2016gcn} is a foundational GNN model that effectively captures graph structure by recursively aggregating feature information from neighboring nodes through a first-order approximation of spectral convolutions. GCN achieves strong performance in tasks such as semi-supervised node classification, link prediction, and community detection. 

\textbf{GCNII} \cite{chen2020gcnii} is an extension of GCN that improves expressiveness by incorporating higher-order graph convolutions while avoiding over-smoothing. It introduces a novel initialization and an implicit regularization mechanism, making it robust for deep graph networks. GCNII excels in tasks involving large and sparse graphs, such as semi-supervised node classification and graph regression, while maintaining stability in deep models.

\textbf{LightGCN} \cite{LightGCN} is a light-weight GNN model optimized for collaborative filtering tasks, specifically in recommender systems. Unlike traditional GCNs, it simplifies the convolution operation by removing unnecessary transformations and focusing solely on neighborhood aggregation. LightGCN delivers state-of-the-art performance in tasks such as link prediction and recommendation while maintaining computational efficiency.

\textbf{GAT} \cite{velivckovic2017gat} enhances traditional GNNs by introducing attention mechanisms to weigh the importance of neighboring nodes when aggregating features. This attention-driven approach makes GAT especially effective in graphs with complex node relationships, offering a robust solution for node classification and link prediction.

\textbf{GATv2} \cite{brody2021gatv2} is an enhanced version of the Graph Attention Network (GAT) that improves the flexibility and expressiveness of attention mechanisms. GATv2 introduces a more generalized attention mechanism, allowing for better handling of noisy graph data and fine-grained aggregation of neighborhood information. It has shown superior performance in node classification and graph classification tasks, especially for graphs with heterogeneous features.


\textbf{GIN}  \cite{xu2018gin} is designed to capture structural distinctions in graphs with maximal expressive power. By applying a sum aggregator followed by a multi-layer perceptron, GIN achieves injective mapping of neighborhood features, allowing it to differentiate between complex graph structures more effectively than other models.

\textbf{GraphSAGE} \cite{hamilton2017graphsage} is a powerful framework that generalizes GNNs by learning node embeddings through sampling and aggregating features from a fixed-size neighborhood. Unlike traditional GNNs, which require access to the entire graph, GraphSAGE supports efficient inductive learning by training on sampled node neighborhoods, enabling it to scale effectively to large graphs and adapt to unseen nodes in dynamic environments. 

\textbf{GraphSAINT} \cite{zeng2019graphsaint} is a scalable GNN model designed to handle large-scale graphs by employing efficient sampling methods. It combines graph sampling with mini-batch training, allowing the model to operate on subgraphs rather than the entire graph, which greatly improves training efficiency. GraphSAINT achieves competitive performance in node classification and graph classification tasks, particularly for large graphs with millions of nodes.

\textbf{Cluster-gcn} \cite{chiang2019cluster-gcn} is a GNN model designed to handle graph clustering tasks by exploiting node-level similarity structures. It divides the graph into multiple subclusters and performs learning on these subgraph clusters to improve performance and scalability. Cluster-gcn has been demonstrated to achieve high accuracy in clustering-based tasks while maintaining computational efficiency in large-scale graphs.

\textbf{SGC} \cite{wu2019sgc} is a streamlined variant of GCN that removes non-linearity between layers, significantly reducing computational complexity. By collapsing multiple layers into a single linear transformation, SGC preserves essential neighborhood information while enhancing scalability, making it particularly suitable for large-scale node classification tasks.

\textbf{SSGC} \cite{zhu2021ssgc} extends the SGC model by incorporating self-supervised learning through contrastive loss, enabling more robust representation learning. This method preserves the efficient linear architecture of SGC while enhancing node embeddings by contrasting positive and negative samples, capturing meaningful graph structures.

\textbf{SIGN} \cite{frasca2020sign}  is designed for large-scale graphs, leveraging a pre-computation strategy to improve efficiency. It processes multiple hops of neighborhood information independently, storing them as separate feature channels. By avoiding recursive message passing during training, SIGN achieves scalability and faster computation, making it well-suited for tasks on massive graph datasets.

\textbf{APPNP} \cite{2019appnp} introduces a novel approximation of personalized PageRank, combining random walk-based methods with graph neural networks. APPNP performs graph-based label propagation with high efficiency, significantly improving node classification performance, particularly in semi-supervised learning settings. The model’s robustness to noisy graphs and its ability to propagate information over distant nodes makes it a powerful tool for graph-based recommendation systems.

\subsection{GU Method Details}
\label{A.GU}
In the context of GU, a range of specialized methods has been developed to address the challenges of selectively forgetting specific nodes, edges, or features in a trained model while preserving the integrity of the remaining graph information. These methods aim to balance effectiveness in unlearning with model efficiency and robustness, ensuring minimal impact on predictive performance for retained data. Below, we provide a detailed overview of these approaches, highlighting their mechanisms and contributions to advancing GU capabilities.

\textbf{GraphEraser} \cite{chen2022graph_eraser} is a GU method designed to efficiently remove specific nodes or edges from a trained GNN model. By utilizing two novel graph partition algorithms and a learning-based aggregation method, GraphEraser identifies and updates only the relevant substructures, enabling precise removal while minimizing computational overhead.

\textbf{GUIDE} \cite{wang2023guide} is an inductive GU framework designed to address the limitations of traditional transductive unlearning approaches like GraphEraser. GUIDE incorporates a balanced graph partitioning mechanism, efficient subgraph repair, and similarity-based aggregation to ensure effective unlearning while maintaining computational efficiency. This approach enables unlearning with low partition costs, preserving model adaptability in inductive graph learning tasks.

\textbf{GraphRevoker} \cite{2024ZhangGraphRevoker} addresses the challenge of GU by balancing efficient unlearning with high model utility. Unlike traditional methods that may fragment essential information during partitioning, GraphRevoker employs property-aware sharding, preserving key structural and attribute information. For final predictions, it integrates sub-GNN models through a contrastive aggregation technique, ensuring coherent model utility while unlearning data.

\textbf{GIF} \cite{wu2023gif} innovatively tackles GU by leveraging a tailored influence function, enhancing both efficiency and precision in unlearning processes. GIF redefines influence to address dependencies among neighboring nodes by introducing a specialized loss term that accounts for structural interactions, which traditional influence functions overlook. This model-agnostic approach enables GIF to estimate parameter adjustments in response to small data perturbations, providing a closed-form solution that facilitates understanding of the unlearning mechanics.

\textbf{ScaleGUN} \cite{scaluGUN} is a scalable framework for certified GU, addressing the inefficiency of traditional methods by integrating approximate graph propagation techniques. It ensures certified guarantees for node feature, edge, and node unlearning scenarios while maintaining bounded model error on embeddings. By balancing efficiency and accuracy, ScaleGUN extends certified unlearning to large-scale graphs without compromising privacy guarantees.

\textbf{CGU} \cite{chien2022cgu} offers the first approximate GU approach with theoretical guarantees, designed to manage varied unlearning requests. CGU emphasizes precise handling of feature mixing during propagation, particularly within SGC. This enables CGU to efficiently balance privacy, complexity, and performance, achieving rapid unlearning with minimal accuracy loss.

\textbf{CEU} \cite{2023WuCEU} introduces an efficient solution for edge unlearning in GNNs by bypassing the high costs of full retraining. CEU uniquely leverages a single-step parameter update on the pre-trained model, effectively erasing edge influence while preserving model integrity. The CEU framework also offers theoretical guarantees under convex loss conditions, achieving notable speedup with model accuracy nearly on par with complete retraining.

\textbf{GST} \cite{pan2023gst_unlearning} offers an innovative approach to GU by leveraging its efficient, stable, and provably resilient framework under feature or topology perturbations. Unlike traditional GNN retraining, GST-based unlearning introduces a nonlinear approximation mechanism that achieves competitive graph classification performance, making GST an advantageous method for privacy-sensitive applications demanding efficient unlearning without sacrificing performance.

\textbf{IDEA} \cite{2024IDEA} represents an outstanding framework in the realm of GU, addressing the critical need for flexible and certified unlearning. It pioneers an approach that accommodates a variety of unlearning requests, transcending the limitations of specialized GNN designs or objectives. IDEA's flexibility is underscored by its ability to provide theoretical guarantees for information removal across diverse GNN architectures, setting a new standard for privacy protection.

\textbf{GNNDelete} \cite{cheng2023gnndelete} addresses key challenges in GU by introducing a model-agnostic, layer-wise operator that effectively removes graph elements. Its core mechanisms, Deleted Edge Consistency and Neighborhood Influence, ensure that deleted edges and nodes are fully excluded from model representations without compromising the influence of neighboring nodes.

\textbf{MEGU} \cite{xkliMEGU2024} introduces a pioneering mutual evolution approach for GU, where prediction accuracy and unlearning effectiveness evolve synergistically within a single training framework. MEGU’s adaptability enables precise unlearning at the feature, node, and edge levels, showing strong superiority in the field of GU.

\textbf{SGU} \cite{} is a pioneering approach that addresses the challenges of gradient-driven node entanglement and scalability in GU. By integrating Node Influence Maximization, SGU identifies the highly influenced nodes through a decoupled influence propagation model, offering a scalable and plug-and-play solution.

\textbf{D2DGN} \cite{2024D2DGN} revolutionizes GU through knowledge distillation. It adeptly addresses the complexities of removing specific elements from GNNs by dividing and marking graph knowledge for retention and deletion. D2DGN stands out for its efficiency, effectively eliminating the influence of deleted elements while preserving retained knowledge, and its superior performance in both edge and node unlearning tasks across real-world datasets.


\textbf{GUKD} \cite{2023GUKD} harnesses the power of knowledge distillation, emerging as an innovative solution for class unlearning in GNN. It distinctively pairs a deep GNN model with a shallow student network, facilitating the transfer of retained knowledge and enabling targeted forgetting. GUKD stands out with its exceptional performance, thereby setting a new benchmark for efficient and effective GU.

\textbf{UtU} \cite{Tan2024UtU} represents a paradigm shift in edge unlearning for GNNs, offering a streamlined solution that eschews the pitfalls of over-forgetting associated with traditional methods. This innovative technique simplifies the unlearning process by directly unlinking specified edges from the graph. UtU's elegance lies in its ability to safeguard privacy with minimal computational overhead, aligning closely with the performance of a freshly retrained model while ensuring the integrity of the remaining graph structure.

\textbf{Projector} \cite{cong2023projector} stands out by projecting the pre-trained model's weight parameters onto a subspace unrelated to the features of unlearning nodes, effectively bypassing node dependency issues. This method ensures a perfect data removal, where the unlearned model parameters are devoid of any information concerning the unlearned nodes, a guarantee inherent in its algorithmic design.

\subsection{Evaluation Metric Details}
\label{A.Evaluation Metrics}

Our OpenGU includes three downstream tasks in total: node classification, link prediction, and graph classification. These three downstream tasks are evaluated using F1-score, ROC-AUC, and accuracy Acc, respectively.

\textbf{F1-score}, a metric used to evaluate classification problems, is the harmonic mean of precision and recall. In node classification tasks, there may be an imbalance in categories, that is, the number of samples in some categories is much larger than that in other categories. In this case, using accuracy alone may lead to misjudgment of model performance. F1-score can more comprehensively reflect the performance of the model in each category by considering both precision and recall, especially when dealing with data with imbalanced categories.

\textbf{ROC-AUC}, the area under the ROC curve, is used to evaluate the performance of a classification model. AUC values range from 0 to 1, with larger values indicating better performance. In link prediction tasks, the AUC value measures the model's ability to predict whether there is a link between node pairs. A higher AUC value indicates that the model is more effective in distinguishing nodes that actually have links from node pairs that do not.

\textbf{Accuracy}, one of the basic indicators for measuring model performance. In graph classification tasks, accuracy is the most intuitive evaluation indicator and is easy to understand and calculate.

\textbf{Precision}, a crucial indicator for model performance assessment. In some tasks, it shows the proportion of correctly predicted positive instances among all predicted positives. It is a significant metric that helps to understand the exactness of the model's predictions.

\subsection{Attack Details}
\label{A.Attack}

We implemented two attack strategies, Membership Inference Attack and Poison Attack, and evaluated them respectively.

\textbf{Membership Inference Attack}, a privacy attack technique against machine learning models in which the attacker attempts to determine whether a specific data record was previously used to train the machine learning model. We use Membership Inference Attack to gauge the efficacy of unlearning by quantifying the probability ratio of unlearned part presence before and after the unlearning process.

\textbf{Poisoning Attack}, a security threat against machine learning models. Its main purpose is to disrupt the model training process by contaminating the training data during the model training phase. Add heterogeneous edges to the original data as negative samples, and judge the quality of the forgetting effect by comparing the effects of the model before and after forgetting the heterogeneous edges.

\subsection{Experimental Setting Details}
\label{A.set}
All experiments were conducted on a system equipped with an NVIDIA A100 80GB PCIe GPU and an Intel(R) Xeon(R) Gold 6240 CPU @ 2.60GHz, with CUDA Version 12.4 enabled. The software environment was set up with Python 3.8.0 and PyTorch 2.2.0 to ensure optimal compatibility and performance for all GU algorithms. Additionally, hyperparameters for each GU algorithm were configured based on conclusions drawn from prior research to provide consistent and reliable results.

\end{document}